%% file: main.tex
\documentclass[10pt,twocolumn,letterpaper]{article}

\usepackage[final,algorithms]{wacv}

\definecolor{wacvblue}{rgb}{0.21,0.49,0.74}
\usepackage[hyphens]{url}
\usepackage[pagebackref,breaklinks,colorlinks,allcolors=wacvblue]{hyperref}

\def\wacvPaperID{1856}
\def\confName{WACV}
\def\confYear{2026}

\usepackage{microtype}
\usepackage[T1]{fontenc}
\usepackage{graphicx}
\usepackage{amsmath,amssymb,amsfonts}
\usepackage{booktabs}
\usepackage{multirow}
\usepackage{xcolor}
\usepackage{pifont}
\usepackage{float}

\newcommand{\cmark}{\ding{51}}
\newcommand{\xmark}{\ding{55}}
\newcommand{\spm}[1]{{\scriptsize$\,\pm\,$#1}}

\begin{document}

\title{BTI-Net: Bidirectional Decoder-Level Task Interaction via\\
Uncertainty-Aware Gating for Multi-Task Medical Image Analysis}

\author{
Abdullah Al Shafi$^{1\dagger}$\quad
Md Kawsar Mahmud Khan Zunayed$^{1\dagger}$\quad\\
Safin Ahmmed$^{1}$\quad
Sk Imran Hossain$^{1}$\quad
Engelbert Mephu Nguifo$^{2}$
\\[4pt]
{\small
$^{1}$Khulna University of Engineering \& Technology, Bangladesh
\quad
$^{2}$University Clermont Auvergne, France
}
}

\maketitle
\renewcommand\thefootnote{}\footnotetext{$^\dagger$Equal contribution.}\renewcommand\thefootnote{\arabic{footnote}}

\begin{abstract}
Jointly learning to segment and classify medical images demands
cross-task synergy, yet encoder-sharing architectures limit decoder
reconstruction to task-private representations, permanently discarding
the boundary cues and semantic priors each branch could supply to the
other.
This work introduces BTI-Net, which
establishes bidirectional communication at every decoder level through
two parallel pathways via Task Interaction Modules (TIM). Spatial boundary context is gated into the
classification branch, while global semantic priors multiplicatively
modulate the decoder, with refined features propagating progressively
from coarse semantics to fine boundary detail across all four decoder
resolutions.
Since cross-task interaction is not equally reliable for every input,
Uncertainty Proxy Attention (UPA) gates each TIM output per instance
and per level using three signals that capture cross-task alignment,
scene complexity, and prediction confidence, without external
annotations or additional inference passes.
Experiments on three medical benchmarks spanning ultrasound,
dermoscopy, and brain MRI demonstrate
consistent improvements in segmentation IoU and classification
accuracy over both encoder-sharing and decoder-interaction baselines. Ablation confirms adaptive gating
contributes $+$2.36~IoU over fixed bidirectional interaction, and
classification accuracy improves by up to $+$2.26 points over the
strongest multi-task baseline.
UPA's uncertainty proxies serve as reliable single-pass task-failure
signals without the overhead of stochastic sampling.
Code: \url{https://github.com/C-loud-Nine/BTI-Net_MTL}
\end{abstract}

\section{Introduction}
MTL enables a single model to address multiple related tasks jointly,
with the dominant approach coupling a shared encoder to independent
decoder heads.
While efficient, this design restricts all inter-task communication
to a coarse bottleneck, after which the decoders diverge independently.
Segmentation recovers precise boundaries that could disambiguate
classification; classification accumulates global semantic context
that could resolve ambiguous spatial regions.
Both signals are lost once the decoders diverge, a gap that is
especially consequential in medical imaging where datasets are small,
pathology is heterogeneous, and complementary task signals carry real
diagnostic value.

Recent decoder-interaction methods partially address this.
MTI-Net~\cite{vandenhende2020mtinet} shows decoder-level interaction
outperforms encoder-sharing, but propagates information in one
direction with fixed per-sample blending.
DenseMTL~\cite{lopes2023densemtl} adds bidirectionality but applies
uniform blending regardless of scene complexity.
In medical imaging, per-instance adaptability is not optional: the
appropriate interaction strength varies fundamentally across pathology
types, imaging conditions, and lesion morphology.

Addressing both limitations, BTI-Net introduces Task Interaction
Modules (TIM) for bidirectional decoder-level communication, paired
with Uncertainty Proxy Attention (UPA) for per-instance adaptive
gating.
TIM establishes bidirectional communication at every decoder level,
from coarsest semantics to finest boundary detail: spatially aggregated
boundary cues are injected into the classification vector through a
gated residual (Seg$\rightarrow$Clf), while global semantic priors are
multiplicatively broadcast across the decoder map
(Clf$\rightarrow$Seg), with refined features carried forward
progressively across all four resolutions.
UPA gates each TIM output per instance and per level using three
signals that capture cross-task alignment, scene complexity, and
prediction confidence.
Independent sigmoid weights allow both tasks to be simultaneously
enhanced when interaction is reliable; on easy samples the gate
remains conservative, preserving the unperturbed representations.
A difficulty-aware loss aligns the gate with per-sample task
performance, requiring no external annotations and no additional
inference passes.

Evaluated on three publicly available benchmarks spanning ultrasound,
dermoscopy, and brain MRI, BTI-Net advances both segmentation and
classification consistently across modalities.
Ablation confirms that adaptive gating contributes $+$2.36~IoU over
fixed bidirectional interaction, and the complete model improves
classification accuracy by up to $+$2.26 points over the strongest
multi-task baseline.

The main contributions are:
\begin{enumerate}
  \item TIM: bidirectional decoder-level interaction between
    segmentation and classification via channel-wise gated operations,
    with refined spatial features propagated progressively across
    all four decoder resolutions.
  \item UPA: a per-instance, per-level adaptive gate using three
    interpretable uncertainty signals, trained from per-sample
    difficulty scores derived automatically from task performance,
    without external annotations or Bayesian overhead.
  \item Cross-modality validation on three diverse medical benchmarks,
    including, to the best of our knowledge, the first
    decoder-interaction MTL evaluation on BRISC.
\end{enumerate}

\section{Related Work}

\subsection{Multi-Task Learning Architectures}

MTL architectures divide broadly into encoder-sharing and
decoder-interaction approaches.
Encoder-sharing methods~\cite{caruana1997multitask,liu2019endtoend,gao2020mtlnas}
couple a common feature extractor to task-specific heads that diverge
during upsampling, leaving cross-task signals untapped after the shared
bottleneck.

Several works introduce cross-task connections during upsampling.
PAD-Net~\cite{xu2018padnet} distills auxiliary task predictions into
final tasks unidirectionally, establishing the pattern of using
intermediate task outputs to guide final predictions.
MTI-Net~\cite{vandenhende2020mtinet} models interactions at every
backbone scale via distillation units, showing decoder-level
interaction outperforms encoder-sharing; however, the flow is strictly
unidirectional with identical blending applied to every sample.
DenseMTL~\cite{lopes2023densemtl} adds bidirectional pairwise
cross-task attention in decoder exchange blocks, yet applies a fixed
blending weight regardless of scene complexity.
Transformer-based methods such as InvPT~\cite{ye2022invpt} and
TaskPrompter~\cite{ye2023taskprompter} model dependencies through
global self-attention across task tokens; because they operate on
joint token sets rather than explicit directed pathways, they do not
provide the separate per-task gating that characterizes bidirectional
decoder interaction.

Three limitations persist across all prior decoder-interaction methods:
attention costs that grow with spatial resolution make them poorly
suited to small medical datasets; fixed blending ignores per-instance
variability in cross-task reliability; and none has been evaluated on
medical imaging.
BTI-Net addresses all three: TIM's cost scales with channel dimension
rather than spatial resolution, UPA provides per-instance adaptive
gating, and the method is validated across three medical modalities,
as summarized  in Table~\ref{tab:rw_comparison}.

\begin{table}[t]
\centering
\caption{Positioning of BTI-Net against representative MTL
architectures.
\emph{Decoder}: cross-task connections via explicit directed pathways;
\emph{Bidir.}: both tasks update each other at the same resolution;
\emph{Per-inst.}: interaction strength adapts per sample;
\emph{Medical}: validated on medical imaging.}
\label{tab:rw_comparison}
\setlength{\tabcolsep}{3.0pt}
\footnotesize
\begin{tabular}{lcccc}
\toprule
\textbf{Method} & \textbf{Decoder} & \textbf{Bidir.} &
\textbf{Per-inst.} & \textbf{Medical} \\
\midrule
MTAN~\cite{liu2019endtoend}            & \xmark & \xmark & \xmark & \cmark \\
PAD-Net~\cite{xu2018padnet}            & \cmark & \xmark & \xmark & \xmark \\
MTI-Net~\cite{vandenhende2020mtinet}   & \cmark & \xmark & \xmark & \xmark \\
MTANet~\cite{chen2023mtanet}           & \xmark & \xmark & \xmark & \cmark \\
MTL-OCA~\cite{gao2025mtloca}           & \xmark & \xmark & \xmark & \cmark \\
DenseMTL~\cite{lopes2023densemtl}      & \cmark & \cmark & \xmark & \xmark \\
InvPT~\cite{ye2022invpt}               & \cmark & \xmark & \xmark & \xmark \\
DynaShare~\cite{rahimian2023dynashare} & \xmark & \xmark & \cmark & \xmark \\
\midrule
\textbf{BTI-Net}                       & \cmark & \cmark & \cmark & \cmark \\
\bottomrule
\end{tabular}
\end{table}

\subsection{Uncertainty in Multi-Task Learning}

Uncertainty modelling in MTL has focused primarily on task-level loss
balancing.
Kendall et al.~\cite{kendall2018multitask} proposed weighting losses by
homoscedastic task uncertainty, while GradNorm~\cite{chen2018gradnorm}
and PCGrad~\cite{yu2020pcgrad} balance tasks by normalising or
projecting conflicting gradients.
MTANet~\cite{chen2023mtanet} and MTL-OCA~\cite{gao2025mtloca} extend
these ideas with attention-based dynamic weighting.
All these approaches operate at the task level---scaling each task's
loss contribution---and cannot modulate cross-task feature exchange
strength for individual inputs.

Monte Carlo dropout offers per-sample uncertainty estimates but
requires multiple stochastic forward passes, prohibitive for dense
prediction.
The closest approach to instance-level feature adaptation is
DynaShare~\cite{rahimian2023dynashare}, which learns a per-input
gating policy over encoder layers; however, it gates layer execution
rather than cross-task feature exchange, requires a separate policy
network, and has not been applied to dense medical analysis.
UPA gates cross-task enhancement per instance and per decoder level
without additional inference cost.

\subsection{Medical Image Analysis with MTL}

Joint segmentation and classification has been widely studied for
ultrasound~\cite{chen2023mtanet,gao2025mtloca},
dermoscopy~\cite{xie2020mutual}, and brain
MRI~\cite{myronenko2019robust}.
MTANet~\cite{chen2023mtanet} and MTL-OCA~\cite{gao2025mtloca}
incorporate attention and optimal channel selection to improve task
synergy; MFFMT~\cite{wei2024mffmt} introduces adaptive multi-feature
fusion to reduce information-sharing conflicts.
All these methods operate within the encoder-sharing paradigm, apply
fixed interaction strategies to every sample, and have been evaluated
on a single modality, leaving cross-modal generalizability and
decoder-level interaction in medical imaging as open problems.

\section{Method}
\label{sec:method}

\begin{figure*}[t]
\centering
\includegraphics[width=\textwidth]{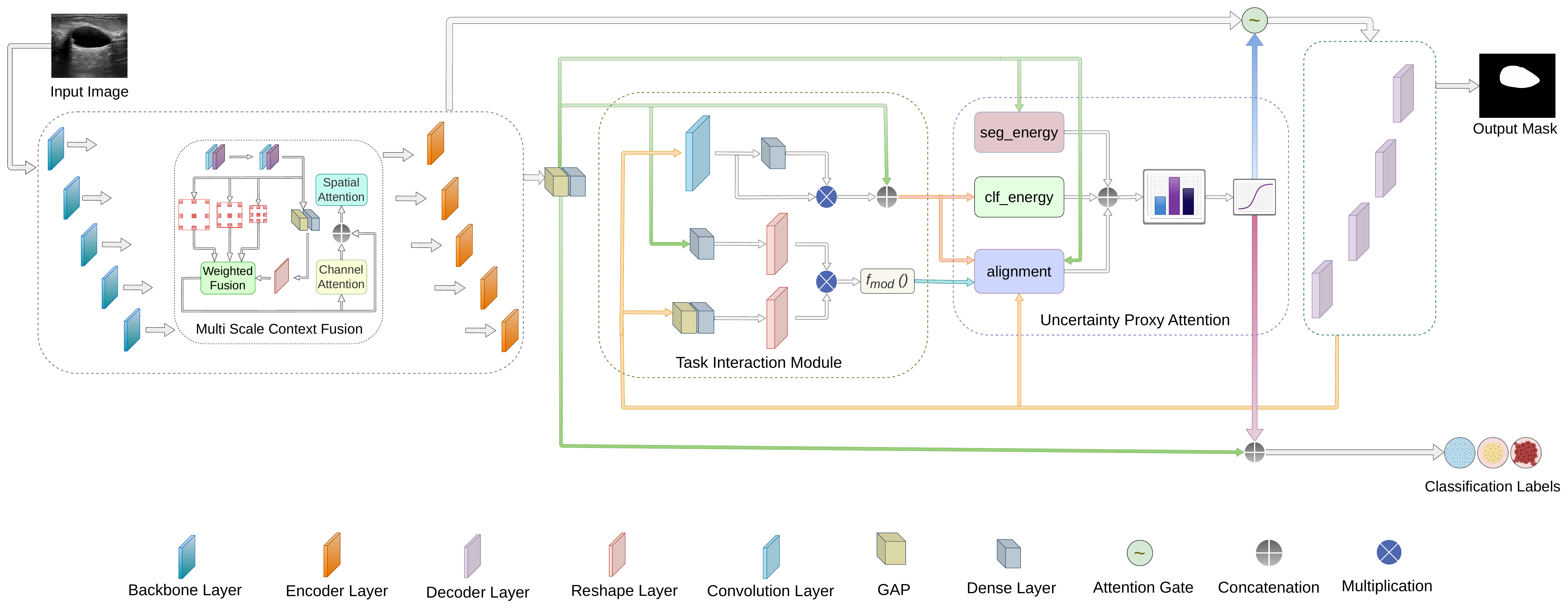}
\caption{Architecture overview.
Multi-scale context fusion augments encoder features at each stage.
TIM establishes bidirectional task interaction at all four decoder levels.
UPA gates the interaction strength per instance using three interpretable
uncertainty proxies.
Dual heads produce the segmentation mask and classification prediction.}
\label{fig:architecture}
\end{figure*}

The proposed framework augments an EfficientNet-B4~\cite{tan2019efficientnet}
encoder with two modules, TIM and UPA, that together enable adaptive
bidirectional task interaction at every decoder level.
Figure~\ref{fig:architecture} shows the complete pipeline.
The encoder comprises five hierarchical stages
($s_1$: 144-ch, $s_2$: 192-ch, $s_3$: 336-ch, $s_4$: 960-ch,
bridge: 1792-ch), each followed by a Multi-Scale Context Fusion (MSCF)
block that applies three parallel dilated separable convolutions
($r\!\in\!\{1,2,4\}$) with squeeze-and-excitation scale
weighting~\cite{chen2018deeplab,hu2018squeeze}.
A four-level decoder ($D_1$--$D_4$, channels 384/192/96/48,
resolutions $14^2$--$112^2$) reconstructs the segmentation mask via
transposed convolutions and attention-gated skip
connections~\cite{oktay2018attention}.
At every decoder level, TIM establishes bidirectional cross-task
communication and UPA gates the resulting enhancement per instance
before the output is upsampled to the next level.
A $224\!\times\!224$ segmentation head and a multi-scale classification
head aggregating 4304-D of encoder and decoder features
(Sec.~\ref{sec:upa}) produce the dual-task predictions.

\subsection{Task Interaction Module}
\label{sec:tim}

Prior decoder-interaction methods either propagate information
unidirectionally or apply fixed per-sample blending; TIM addresses
both by establishing \emph{bidirectional} communication at \emph{every}
decoder level.

At level $\ell\!\in\!\{1,2,3,4\}$, let
$D_\ell\!\in\!\mathbb{R}^{B\times H_\ell\times W_\ell\times C_\ell}$
denote the decoder feature map and
$f_{\mathrm{clf}}^{\ell}\!\in\!\mathbb{R}^{B\times256}$ a level-specific
classification vector.
Each $f_{\mathrm{clf}}^{\ell}$ is initialized by global average pooling (GAP)
and a learned linear projection from the encoder stage whose semantic
abstraction best matches that decoder level:
bridge (1792-D) $\to D_1$, $s_4$ (960-D) $\to D_2$,
$s_3$ (336-D) $\to D_3$, $s_2$ (192-D) $\to D_4$.
This pairing anchors the most abstract representation at the coarsest
decoder level and progressively finer spatial cues at deeper levels.

In the Seg$\rightarrow$Clf direction, spatial boundary context recovered
by the decoder is injected into the classification vector through a
gated residual.
A $1\!\times\!1$ convolution reduces the decoder map to $C_\ell/2$
channels before global average pooling yields a compact spatial summary
$\mathbf{z}_\ell$, which is then projected and gated:
\begin{align}
\mathbf{z}_\ell &=
  \mathrm{GAP}\!\bigl(\mathrm{Conv}_{1\times1}^{C_\ell/2}(D_\ell)\bigr),
  \label{eq:s2c1}\\
\mathbf{c}_\ell &=
  \mathrm{Dense}_{256}\!\bigl(\mathrm{ReLU}(\mathbf{z}_\ell)\bigr),
  \label{eq:s2c2}\\
\mathbf{g}_{\mathrm{clf}} &=
  \sigma\!\bigl(\mathrm{Dense}_{256}(\mathbf{c}_\ell)\bigr),
  \label{eq:s2c3}\\
f_{\mathrm{clf}}^{\ell,\mathrm{enh}} &=
  f_{\mathrm{clf}}^{\ell} +
  \mathbf{g}_{\mathrm{clf}}\odot\mathbf{c}_\ell.
  \label{eq:s2c4}
\end{align}
The element-wise gate $\mathbf{g}_{\mathrm{clf}}\!\in\![0,1]^{256}$
selects which spatial cues to admit per sample, while the residual
in Eq.~\eqref{eq:s2c4} preserves the original classification semantics.

In the complementary Clf$\rightarrow$Seg direction, global semantic
priors from the classification branch modulate the decoder feature map
channel-wise.
The classification vector is projected to the decoder's channel space
and combined with a self-gate derived from the decoder features:
\begin{align}
\mathbf{u}_{\mathrm{clf}} &=
  \sigma\!\bigl(\mathrm{Dense}_{C_\ell}(f_{\mathrm{clf}}^{\ell})\bigr)
  \in\mathbb{R}^{B\times1\times1\times C_\ell},
  \label{eq:c2s1}\\
\mathbf{g}_{\mathrm{seg}} &=
  \sigma\!\bigl(\mathrm{Dense}_{C_\ell}(\mathrm{GAP}(D_\ell))\bigr),
  \label{eq:c2s2}\\
\boldsymbol{\mu}_\ell &=
  \mathbf{1}+\tau\cdot
  \mathbf{g}_{\mathrm{seg}}\odot\mathbf{u}_{\mathrm{clf}},
  \label{eq:c2s3}\\
D_\ell^{\mathrm{enh}} &=
  D_\ell\odot\boldsymbol{\mu}_\ell.
  \label{eq:c2s4}
\end{align}
The self-gate $\mathbf{g}_{\mathrm{seg}}$ adapts the modulation so that
only channels already consistent with the semantic prior are amplified.
The scalar $\tau\!=\!0.7$ bounds the multiplicative factor to
$[1,\,1.7]$ and was selected by grid search over
$\tau\!\in\!\{0.3,\,0.5,\,0.7,\,1.0\}$.

Both pathways receive the \emph{original} pre-interaction features as
input, making them parallel operations that prevent double-counting:
boundary context injected into the classifier does not contaminate the
segmentation modulation.
Each $D_\ell^{\mathrm{final}}$ (produced by UPA, Sec.~\ref{sec:upa})
feeds the transposed convolution constructing $D_{\ell+1}$, so
refinements accumulate from coarse semantic exchange at
$D_1$ ($14\!\times\!14$) to fine boundary detail at
$D_4$ ($112\!\times\!112$).
The classification branch follows a complementary design: each level
draws a fresh encoder-initialized vector and all four UPA-gated outputs
are aggregated at the multi-scale head, fusing cross-task information
from every decoder resolution without cross-level propagation of
intermediate representations.

\subsection{Uncertainty Proxy Attention}
\label{sec:upa}

Cross-task enhancement is not uniformly beneficial: on ambiguous
low-contrast inputs, TIM-enhanced features may carry misleading signals,
making reversion toward the original representations the safer choice.
UPA interpolates between pre-TIM features
$(D_\ell,\,f_{\mathrm{clf}}^{\ell})$ and TIM-enhanced features
$(D_\ell^{\mathrm{enh}},\,f_{\mathrm{clf}}^{\ell,\mathrm{enh}})$
using two independent scalar weights
$w_{\mathrm{seg}},\,w_{\mathrm{clf}}\!\in\![0,1]$ per sample and per
level, produced by a two-layer MLP driven by three
uncertainty signals.

Three signals form the gate descriptor
$\mathbf{u}_\ell = [a_\ell,\,E_{\mathrm{seg}},\,E_{\mathrm{clf}}]
\in\mathbb{R}^3$.
The first, cross-task alignment $a_\ell$, measures whether TIM moves
both branches in consistent latent directions; the spatially averaged
TIM residuals are projected to a shared space and their cosine
similarity computed:
\begin{align}
a_\ell &= \cos(\mathbf{p}_{\mathrm{seg}},\,\mathbf{p}_{\mathrm{clf}}),
  \label{eq:align}\\
\mathbf{p}_{\mathrm{seg}} &=
  \mathbf{W}_{\!\mathrm{s}}\,\mathrm{GAP}(D_\ell^{\mathrm{enh}}\!-\!D_\ell),
  \notag\\
\mathbf{p}_{\mathrm{clf}} &=
  \mathbf{W}_{\!\mathrm{c}}\,
  (f_{\mathrm{clf}}^{\ell,\mathrm{enh}}\!-\!f_{\mathrm{clf}}^{\ell}).
  \notag
\end{align}
The second, segmentation gradient energy $E_{\mathrm{seg}}$, captures
scene complexity via the mean absolute spatial gradient of the pre-TIM
decoder map, where high values indicate intrinsically harder
segmentation:
\begin{equation}
E_{\mathrm{seg}} =
  \frac{1}{C_\ell H_\ell W_\ell}
  \sum_{c,i,j}\!
  \bigl(|D_\ell^{(i+1,j)}\!-\!D_\ell^{(i,j)}|
       +|D_\ell^{(i,j+1)}\!-\!D_\ell^{(i,j)}|\bigr).
\label{eq:seg_energy}
\end{equation}
The third, classification activation spread $E_{\mathrm{clf}}$,
distinguishes confident peaked distributions from flat uncertain ones
through the log-variance of the enhanced classification vector:
\begin{equation}
E_{\mathrm{clf}} =
  \log\!\Bigl(1 +
  \tfrac{1}{256}{\textstyle\sum_{k=1}^{256}}
  \bigl(f_{\mathrm{clf}}^{\ell,\mathrm{enh}}[k]
       -\bar{f}_{\mathrm{clf}}^{\ell,\mathrm{enh}}\bigr)^2\Bigr).
\label{eq:clf_energy}
\end{equation}
Together, $a_\ell$, $E_{\mathrm{seg}}$, and $E_{\mathrm{clf}}$ span
interaction coherence, scene complexity, and prediction confidence,
covering three complementary uncertainty axes without additional
inference passes or external annotations.

The descriptor $\mathbf{u}_\ell$ is mapped by a two-layer MLP to two
logits, and sigmoid activation yields independent gate weights:
\begin{equation}
\bigl[w_{\mathrm{seg}},\,w_{\mathrm{clf}}\bigr] =
  \sigma\!\bigl(
  \mathrm{Dense}_2(\mathrm{ReLU}(\mathrm{Dense}_{32}(\mathbf{u}_\ell)))
  \bigr).
\label{eq:gate}
\end{equation}
Sigmoid is used rather than softmax because the zero-sum constraint
$w_{\mathrm{seg}}+w_{\mathrm{clf}}=1$ suppresses one task whenever
the other is enhanced, even on easy samples where both branches are
reliable.
Independent sigmoid weights allow both to approach~1 simultaneously,
making UPA a confidence-based interpolator rather than a task selector.
The final features are produced by residual interpolation:
\begin{align}
D_\ell^{\mathrm{final}} &=
  D_\ell + w_{\mathrm{seg}}\,
  (D_\ell^{\mathrm{enh}}\!-\!D_\ell),
  \label{eq:upa_seg}\\
f_{\mathrm{clf}}^{\ell,\mathrm{final}} &=
  f_{\mathrm{clf}}^{\ell} + w_{\mathrm{clf}}\,
  (f_{\mathrm{clf}}^{\ell,\mathrm{enh}}\!-\!f_{\mathrm{clf}}^{\ell}).
  \label{eq:upa_clf}
\end{align}

A difficulty-aware auxiliary loss trains $w_{\mathrm{seg}}$ and
$w_{\mathrm{clf}}$ directly on per-sample task performance without
external annotations.
Per-sample soft-IoU values are min-max normalized  within the mini-batch
and inverted so that the hardest sample receives a target of~1,
encouraging the gate to be sensitive to sample difficulty:
\begin{equation}
t_i^{\mathrm{seg}} =
  1 - \frac{\mathrm{IoU}_i - \mathrm{IoU}_{\min}}
           {\mathrm{IoU}_{\max} - \mathrm{IoU}_{\min} + \varepsilon}.
\label{eq:target}
\end{equation}
On ambiguous inputs where cross-task alignment $a_\ell$ is
low, the gate simultaneously receives a signal to suppress
the interaction; the difficulty target and the alignment
signal thus act complementarily, with the net realized
enhancement reflecting both.
An analogous target $t_i^{\mathrm{clf}}$ is derived from inverted
min-max prediction confidence.
The gate is trained by minimizing BCE between each target and its
gate weight, averaged over all four UPA levels as
$\mathcal{L}_{\mathrm{gate}}$, yielding the complete training objective:
\begin{equation}
\mathcal{L} =
  \mathcal{L}_{\mathrm{seg}} +
  \mathcal{L}_{\mathrm{clf}} +
  \mathcal{L}_{\mathrm{gate}}.
\label{eq:loss}
\end{equation}
In practice, $\mathcal{L}_{\mathrm{gate}}$ is activated only in
Stage~2 after the main network has converged; during Stage~1,
$\lambda_{\mathrm{gate}}\!=\!0$ and the model trains under
$\mathcal{L}_{\mathrm{seg}} + \mathcal{L}_{\mathrm{clf}}$ only.

After all four UPA levels, the classification branch aggregates the
encoder's multi-scale features by global average pooling of $s_2$,
$s_3$, $s_4$, and bridge (3280-D total), concatenated with the four
256-D $f_{\mathrm{clf}}^{\ell,\mathrm{final}}$ vectors, yielding a
4304-D representation that fuses hierarchical encoder semantics with
task-refined features across every decoder resolution.

\section{Experiments and Results}
\label{sec:experiments}

\subsection{Datasets and Implementation}
\label{sec:implementation}

Three publicly available benchmarks spanning distinct imaging modalities
are used for evaluation.
BUSI~\cite{al2020dataset} comprises 780 breast ultrasound images across
three clinical classes from 600 patients; normal cases carry empty
segmentation masks and are excluded from IoU computation but retained
for classification.
HAM10000~\cite{tschandl2018ham} is a 10{,}015-image dermoscopic
collection spanning seven diagnostic categories of pigmented skin
lesions; the full dataset with expert-verified segmentation masks
provided by Tschandl et al.~\cite{tschandl2020human} is used.
BRISC~\cite{brisc2026} is a recently released brain MRI benchmark of
6{,}000 T1-weighted scans annotated with tumor segmentation masks and
four-class tumor-type labels; splits are patient-wise with
stratification across classes.
Results are reported as mean and standard deviation over three
independent seeds.

All models share an EfficientNet-B4 encoder pre-trained on ImageNet.
Standard augmentation is applied during training.
Segmentation uses Focal Tversky loss~\cite{abraham2019focal}
($\gamma\!=\!0.75$) and classification uses Focal Cross-Entropy
($\gamma\!=\!2.0$).
Training proceeds in two stages: Stage~1 optimizes the full network
under $\mathcal{L}_{\mathrm{seg}}$ and $\mathcal{L}_{\mathrm{clf}}$;
Stage~2 freezes all parameters except the four UPA gate networks and
introduces $\mathcal{L}_{\mathrm{gate}}$, ensuring the gate is
supervised by a stable predictor.
Gate difficulty targets are derived from per-sample soft-IoU and
prediction confidence, min-max normalized within the mini-batch;
min-max is preferred over rank normalization because it preserves the
magnitude of performance gaps, producing more discriminative gate
supervision.
The gate loss is applied with a stop-gradient on the task predictions,
preventing the gate from gaming the soft-IoU score.
IoU is computed over foreground lesion pixels only; Dice scores are
provided in the supplementary material.
All baselines are reproduced under identical conditions; full
hyperparameter settings and additional uncertainty analyses are
provided in the supplementary material.
Single-task segmentation baselines are evaluated without a
classification head.

\subsection{Ablation Study}
\label{sec:ablation}

All ablations use BUSI under the protocol of Sec.~\ref{sec:implementation} unless noted; uncertainty proxy evaluations extend to HAM10000 and BRISC to verify that observed proxy behavior is not modality-specific.

\begin{figure}[t]
\centering
\includegraphics[width=\columnwidth]{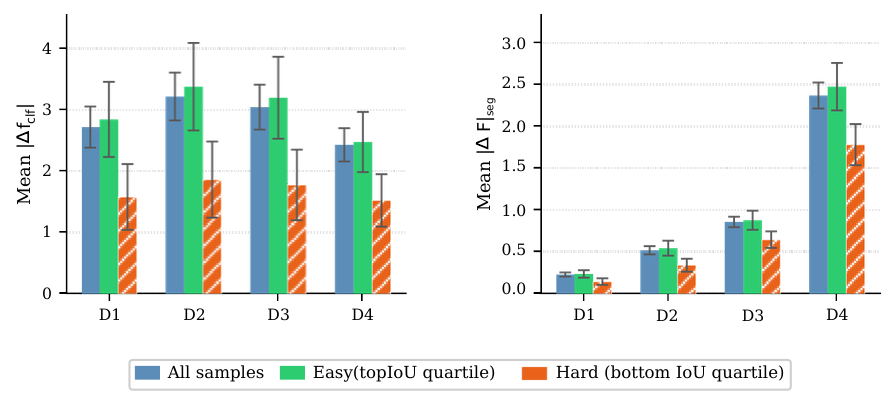}
\caption{TIM feature shift magnitudes on BUSI by sample difficulty
(top/bottom IoU quartile).
\emph{Left} (Seg$\rightarrow$Clf): easy samples receive consistently
larger enhancement, peaking at $D_2$.
\emph{Right} (Clf$\rightarrow$Seg): both groups grow toward $D_4$,
with easy samples maintaining larger absolute enhancement throughout.}
\label{fig:busi_tim_magnitudes}
\end{figure}

Figure~\ref{fig:busi_tim_magnitudes} decomposes TIM's feature modifications $|\Delta f_{\mathrm{clf}}|$ and $|\Delta F_{\mathrm{seg}}|$ by decoder level and sample difficulty (top vs.\ bottom IoU quartile).
In the Seg$\rightarrow$Clf direction, easy samples receive
consistently larger enhancement than hard samples across all levels, peaking at $D_2$, consistent with globally coherent cases benefiting
more from semantic-level cross-task agreement.
In the Clf$\rightarrow$Seg direction, both groups grow monotonically
from $D_1$ to $D_4$, but easy samples receive greater enhancement
at every level and the gap widens at $D_4$, suggesting that
boundary-clear inputs exploit class priors more effectively at fine
resolutions while ambiguous hard samples remain gating-suppressed.
This level-dependent allocation motivates a per-instance, per-level gate rather than a fixed dataset-wide blending coefficient.

UPA's gate signals are evaluated as deterministic, single-pass failure proxies against 30-pass Monte Carlo Dropout.
A sample is labelled a segmentation failure if IoU~$<0.5$ and a classification failure if the predicted class is incorrect; these labels score the already-computed gate values and are never used to select or tune them.
Table~\ref{tab:segw_main_expanded} reports AUC-ROC and AUPRC for $w_{\mathrm{seg}}$ at every decoder level without any level selection.
On HAM10000, $w_{\mathrm{seg}}$ exceeds MC Dropout on AUC-ROC
at all four levels; AUPRC remains below the MC reference,
reflecting the near-zero segmentation failure rate (1.0\%)
that limits precision-recall discriminability.
On BRISC the signal is informative mainly at $D_3$, consistent with the level-dependent enhancement patterns.
The per-modality variation confirms that UPA's reliability signal is level and modality-dependent rather than a single uniformly dominant proxy.

\begin{table}[t]
\centering
\caption{Segmentation-failure detection: AUC-ROC and AUPRC of
$w_{\mathrm{seg}}$ vs.\ 30-pass MC Dropout.
\textbf{Bold}: beats MC Dropout at that level.}
\label{tab:segw_main_expanded}
\footnotesize
\setlength{\tabcolsep}{1.5pt}
\begin{tabular}{lcccccccccc}
\toprule
 & \multicolumn{2}{c}{\textbf{MC (30$\times$)}}
 & \multicolumn{2}{c}{\textbf{D1}}
 & \multicolumn{2}{c}{\textbf{D2}}
 & \multicolumn{2}{c}{\textbf{D3}}
 & \multicolumn{2}{c}{\textbf{D4}} \\
\cmidrule(lr){2-3}\cmidrule(lr){4-5}\cmidrule(lr){6-7}
\cmidrule(lr){8-9}\cmidrule(lr){10-11}
\textbf{Dataset} & AUC & PR & AUC & PR & AUC & PR & AUC & PR & AUC & PR \\
\midrule
BUSI
  & 0.751 & 0.431
  & \textbf{0.767} & \textbf{0.512}
  & \textbf{0.824} & \textbf{0.602}
  & 0.682 & 0.333
  & \textbf{0.785} & \textbf{0.624} \\
BRISC
  & 0.702 & 0.238
  & 0.282 & 0.081
  & 0.482 & \textbf{0.247}
  & \textbf{0.834} & \textbf{0.412}
  & 0.518 & 0.141 \\
HAM
  & 0.891 & 0.151
  & \textbf{0.901} & 0.056
  & \textbf{0.918} & 0.068
  & \textbf{0.934} & 0.079
  & \textbf{0.926} & 0.104 \\
\bottomrule
\end{tabular}
\end{table}

Table~\ref{tab:clf_summary_expanded} compares MC Dropout, the parameter-free maximum-softmax-confidence baseline ($1\!-\!\mathrm{conf}$), and the most discriminative UPA signal per dataset.
AUPRC is the more informative metric at low failure rates and is reported alongside AUC-ROC.
UPA surpasses MC Dropout on AUPRC across all three datasets.
On BUSI, $w_{\mathrm{clf}}$ at $D_4$ achieves 0.850 AUC, exceeding both MC Dropout (0.728) and the confidence baseline (0.834).
On BRISC and HAM10000 the confidence baseline remains stronger (0.920 and 0.834); UPA trails it (0.753 and 0.766) while still exceeding MC Dropout, whose BRISC AUC (0.397) falls below chance due to a known dropout failure mode under class imbalance.
Classification failure detection is not the primary claim; the result positions UPA's gate as a useful secondary signal with no additional inference cost.

\begin{table}[t]
\centering
\caption{Classification-failure detection: AUC-ROC and AUPRC.
UPA exceeds MC Dropout on AUPRC across all three datasets.
1-conf achieves higher AUPRC on BRISC and HAM where softmax
confidence is well-calibrated; UPA exceeds 1-conf on both AUC-ROC and AUPRC for BUSI.}
\label{tab:clf_summary_expanded}
\footnotesize
\setlength{\tabcolsep}{4.5pt}
\begin{tabular}{lccccclcc}
\toprule
 & \multicolumn{2}{c}{\textbf{MC (30$\times$)}}
 & \multicolumn{2}{c}{\textbf{1$-$conf}}
 & \multicolumn{3}{c}{\textbf{Best UPA Signal}} \\
\cmidrule(lr){2-3}\cmidrule(lr){4-5}\cmidrule(lr){6-8}
\textbf{Dataset} & AUC & PR & AUC & PR & Signal & AUC & PR \\
\midrule
BUSI  & 0.728 & 0.264 & 0.834 & 0.529
  & $w_{\mathrm{clf}}$ (D4) & \textbf{0.850} & \textbf{0.543} \\
BRISC & 0.397 & 0.013 & \textbf{0.920} & \textbf{0.359}
  & $-\delta$ (D1) & 0.753 & 0.098 \\
HAM   & 0.575 & 0.324 & \textbf{0.834} & \textbf{0.561}
  & $w_{\mathrm{clf}}$ (D1) & 0.766 & 0.394 \\
\bottomrule
\end{tabular}
\end{table}

\begin{figure}[t]
\centering
\includegraphics[width=\columnwidth]{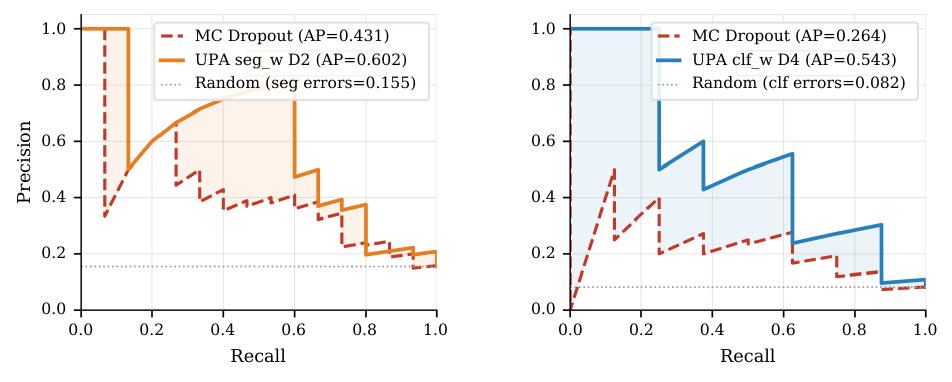}
\caption{Precision-recall curves for failure detection on BUSI.
UPA achieves AP~$=$~0.602 for segmentation ($w_{\mathrm{seg}}$,
$D_2$) and AP~$=$~0.543 for classification ($w_{\mathrm{clf}}$,
$D_4$), against MC Dropout baselines of 0.431 and 0.264.}
\label{fig:busi_roc}
\end{figure}

\begin{figure}[t]
\centering
\includegraphics[width=\columnwidth]{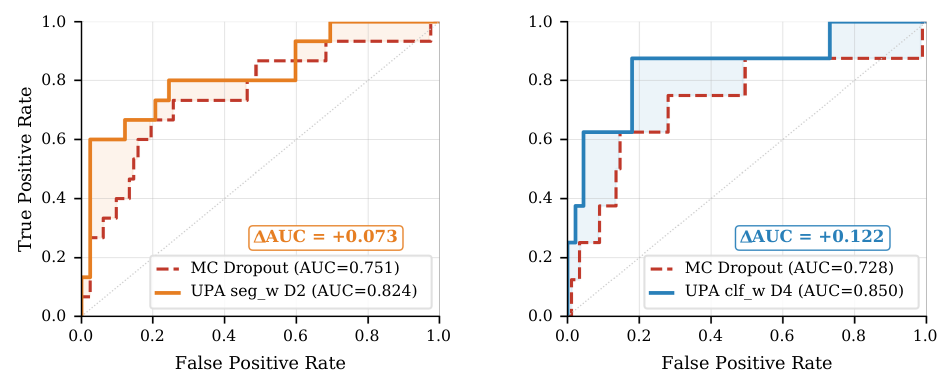}
\caption{ROC curves for failure detection on BUSI across decoder levels
$D_1$--$D_4$. $\Delta$AUC~$=$~$+$0.073 for segmentation ($D_2$) and
$+$0.122 for classification ($D_4$) over MC Dropout.}
\label{fig:busi_scale_analysis}
\end{figure}

Figure~\ref{fig:busi_roc} shows precision-recall curves on BUSI:
$w_{\mathrm{seg}}$ at $D_2$ achieves AP~$=$~0.602 against
MC Dropout's 0.431, and $w_{\mathrm{clf}}$ at $D_4$ achieves
AP~$=$~0.543 against 0.264.
Figure~\ref{fig:busi_scale_analysis} shows the corresponding
ROC curves: $w_{\mathrm{seg}}$ at $D_2$ improves AUC by
$+$0.073 over MC Dropout for segmentation, and $w_{\mathrm{clf}}$
at $D_4$ improves by $+$0.122 for classification.

Table~\ref{tab:ablation_components} isolates each architectural component.
MSCF alone provides $+2.52$~IoU through multi-granularity encoder context.
TIM alone yields $+1.77$~IoU; its modest isolated gain reflects that cross-task interaction requires sufficiently rich encoder representations; paired with MSCF (row~4), the combined gain grows to $+4.71$~IoU, confirming that multi-scale feature quality amplifies bidirectional interaction rather than the contributions being independent.
Adding UPA to MSCF$+$TIM yields the full $+7.07$~IoU, with UPA's isolated contribution of $+1.28$ growing to $+2.36$ in the complete context, consistent with richer features providing more informative gate signals.
The classification accuracy trajectory (row~4 vs.\ row~5) further reveals that UPA specifically benefits classification---TIM$+$UPA achieves higher Acc than MSCF$+$TIM despite lower IoU, because UPA adaptively suppresses low-confidence segmentation priors from contaminating the classification branch.

\begin{table}[t]
\centering
\caption{Component ablation on BUSI.
Mean\,$\pm$\,std over three seeds.}
\label{tab:ablation_components}
\footnotesize
\setlength{\tabcolsep}{3.5pt}
\begin{tabular}{ccc cc}
\toprule
\multicolumn{3}{c}{\textbf{Components}} &
  \multicolumn{2}{c}{\textbf{Results}} \\
\cmidrule(r){1-3}\cmidrule(l){4-5}
MSCF & TIM & UPA & Seg.\ IoU & Clf.\ Acc \\
\midrule
\xmark & \xmark & \xmark & 67.43\spm{1.21} & 84.62\spm{1.08} \\
\cmark & \xmark & \xmark & 69.95\spm{1.04} & 88.89\spm{0.83} \\
\xmark & \cmark & \xmark & 69.20\spm{1.13} & 86.32\spm{0.94} \\
\cmark & \cmark & \xmark & 72.14\spm{0.72} & 90.84\spm{0.58} \\
\xmark & \cmark & \cmark & 70.48\spm{0.91} & 91.47\spm{0.62} \\
\cmark & \cmark & \cmark & \textbf{74.50\spm{0.30}} & \textbf{93.16\spm{0.80}} \\
\bottomrule
\end{tabular}
\end{table}

Replacing the independent sigmoid gates (Eq.~\eqref{eq:gate}) with a softmax constraint creates a zero-sum constraint that penalizes easy samples where both tasks should be enhanced simultaneously.
Removing the difficulty-aware gate regularization (Eq.~\eqref{eq:loss}) drops a further $-0.53$~IoU, indicating that direct supervision toward observed sample difficulty is necessary for the gate to learn a meaningful difficulty ranking rather than deferring entirely to the task losses.

\subsection{Comparison with Prior Work}
\label{sec:comparison}

Comparisons are structured across two complementary tables.
Table~\ref{tab:seg_results} evaluates segmentation performance under
the most favorable conditions for each method: pure single-task
segmentation training for architectures without a native multi-task
design, and joint training for dedicated MTL methods.
Table~\ref{tab:mtl_results} restricts to methods that produce both
outputs simultaneously.

\begin{table*}[t]
\centering
\caption{Joint segmentation and classification results
(mean\,$\pm$\,std over three random seeds) for methods that produce
both outputs simultaneously.
\textbf{Bold}: best per metric--dataset pair.}
\label{tab:mtl_results}
\setlength{\tabcolsep}{5pt}
\renewcommand{\arraystretch}{1.0}
\footnotesize
\begin{tabular}{lc cc @{\hspace{8pt}} cc @{\hspace{8pt}} cc}
\toprule
& & \multicolumn{2}{c}{\textbf{BUSI}}
  & \multicolumn{2}{c}{\textbf{HAM10000}}
  & \multicolumn{2}{c}{\textbf{BRISC}} \\
\cmidrule(lr){3-4}\cmidrule(lr){5-6}\cmidrule(lr){7-8}
\textbf{Method} & \textbf{Year}
  & IoU\,$\uparrow$ & Acc\,$\uparrow$
  & IoU\,$\uparrow$ & Acc\,$\uparrow$
  & IoU\,$\uparrow$ & Acc\,$\uparrow$ \\
\midrule
MTAN~\cite{liu2019endtoend}
  & 2019
  & 68.41\spm{1.12} & 87.84\spm{0.89}
  & 87.24\spm{0.84} & 85.73\spm{0.79}
  & 68.84\spm{0.87} & 93.42\spm{0.67} \\
MTANet~\cite{chen2023mtanet}
  & 2024
  & 71.24\spm{0.74} & 90.47\spm{0.63}
  & 87.12\spm{0.63} & 86.41\spm{0.64}
  & 71.84\spm{0.67} & 95.84\spm{0.52} \\
MTL-OCA~\cite{gao2025mtloca}
  & 2025
  & 72.14\spm{0.61} & 91.23\spm{0.49}
  & 86.84\spm{0.68} & 86.88\spm{0.52}
  & 71.24\spm{0.58} & 96.21\spm{0.36} \\
MTI-Net~\cite{vandenhende2020mtinet}
  & 2020
  & 73.06\spm{0.55} & 90.84\spm{0.53}
  & 87.84\spm{0.44} & 86.73\spm{0.49}
  & 73.84\spm{0.46} & 95.96\spm{0.37} \\
DenseMTL~\cite{lopes2023densemtl}
  & 2023
  & 72.83\spm{0.58} & 91.47\spm{0.51}
  & 88.14\spm{0.53} & 87.12\spm{0.43}
  & 74.61\spm{0.47} & 96.84\spm{0.32} \\
\midrule
\textbf{BTI-Net (Ours)}
  & 2026
  & \textbf{74.50\spm{0.30}} & \textbf{93.16\spm{0.80}}
  & \textbf{89.80\spm{0.20}} & \textbf{87.84\spm{0.58}}
  & \textbf{76.74\spm{0.52}} & \textbf{99.10\spm{0.20}} \\
\bottomrule
\end{tabular}
\end{table*}

\begin{table}[t]
\centering
\caption{Segmentation IoU (mean\,$\pm$\,std over three random seeds).
$\dagger$: evaluated in single-task segmentation mode.
\textbf{Bold}: best per dataset.}
\label{tab:seg_results}
\setlength{\tabcolsep}{3.5pt}
\renewcommand{\arraystretch}{1.0}
\footnotesize
\begin{tabular}{lcccc}
\toprule
\textbf{Method} & \textbf{Year}
  & \textbf{BUSI} & \textbf{HAM10000} & \textbf{BRISC} \\
\midrule
\multicolumn{5}{l}{\textit{CNN-based segmentation}} \\
\midrule
U-Net~\cite{ronneberger2015unet}$^{\dagger}$
  & 2015
  & 66.42\spm{1.12} & 80.12\spm{0.94} & 65.84\spm{0.87} \\
Att-U-Net~\cite{oktay2018attention}$^{\dagger}$
  & 2018
  & 67.89\spm{0.89} & 81.84\spm{0.83} & 68.12\spm{0.79} \\
UNet++~\cite{zhou2018unetpp}$^{\dagger}$
  & 2020
  & 69.12\spm{0.96} & 84.73\spm{0.84} & 71.84\spm{0.81} \\
\midrule
\multicolumn{5}{l}{\textit{Transformer-based segmentation}} \\
\midrule
TransUNet~\cite{wang2024transunet}$^{\dagger}$
  & 2024
  & 70.51\spm{0.71} & 83.84\spm{0.68} & 71.28\spm{0.74} \\
Swin-UNet~\cite{cao2022swin}$^{\dagger}$
  & 2022
  & 71.24\spm{0.63} & 85.14\spm{0.46} & 73.42\spm{0.52} \\
MISSFormer~\cite{huang2023missformer}$^{\dagger}$
  & 2023
  & 72.64\spm{0.54} & 86.82\spm{0.49} & 74.83\spm{0.51} \\
\midrule
\textbf{BTI-Net (Ours)}
  & 2026
  & \textbf{74.50\spm{0.30}}
  & \textbf{89.80\spm{0.20}}
  & \textbf{76.74\spm{0.52}} \\
\bottomrule
\end{tabular}
\end{table}

Table~\ref{tab:seg_results} shows that BTI-Net achieves the highest
segmentation IoU on all three benchmarks.
On BUSI and BRISC, margins over MISSFormer are $+$1.86 and $+$1.91,
reflecting moderate benefit from cross-task boundary priors on
ultrasound and brain MRI.
On HAM10000 the gain is $+$2.98, consistent across both tables: the
seven-class dermoscopy labels encode lesion-type priors that are
uniquely informative for spatial boundary discrimination between
morphologically similar lesion categories---priors that single-task
models cannot access regardless of architectural capacity.

Table~\ref{tab:mtl_results} reports joint performance across all six
metric--dataset pairs, with BTI-Net leading on each.
The progression within the MTL group is architecturally coherent but
not uniform, and reveals dataset-dependent behavior that is
informative in its own right.
On BUSI and BRISC, encoder-sharing methods underperform
single-task transformers on segmentation IoU, confirming that the
encoder-sharing penalty outweighs task-label benefits when data are
limited or domain mismatch is present.
On HAM10000, MTAN approaches and slightly exceeds MISSFormer's
single-task segmentation IoU ($+$0.42), reflecting the partial
compensation that rich seven-class labels provide to shared encoder
representations when training data are abundant; yet MTAN's
classification accuracy on HAM10000 remains the weakest among all
MTL methods, indicating that the segmentation benefit does not
transfer symmetrically to the classification branch.
This asymmetry also produces a crossover: MTANet outperforms MTAN on
BUSI (where medical-specific ultrasound attention dominates) but falls
behind MTAN on HAM10000 (where data volume and label richness favor
general encoder sharing).
Decoder-interaction methods (MTI-Net, DenseMTL) exceed single-task
transformers across all three datasets, confirming that injecting
classification priors during spatial reconstruction is the dominant
driver of improvement; a second crossover appears here, with
DenseMTL trailing MTI-Net on small BUSI ($-$0.23 IoU) but surpassing
it on HAM10000 and BRISC, where the larger datasets support the
richer pairwise cross-task attention.
BTI-Net extends this further: against MTI-Net, segmentation IoU
improves by $+$1.44, $+$1.96, and $+$2.90 on BUSI, HAM10000, and
BRISC, while against DenseMTL the gains are $+$1.67, $+$1.66, and
$+$2.13, attributable to per-instance adaptive gating over fixed
bidirectional blending.
On BRISC, BTI-Net reaches 99.10\% classification accuracy
($+$2.26 over DenseMTL), demonstrating that decoder-level cross-task
gating extracts strong discriminative signals for the four-class brain
tumor task even when segmentation margins are narrower.

\begin{figure*}[t]
\centering
\includegraphics[width=0.85\linewidth]{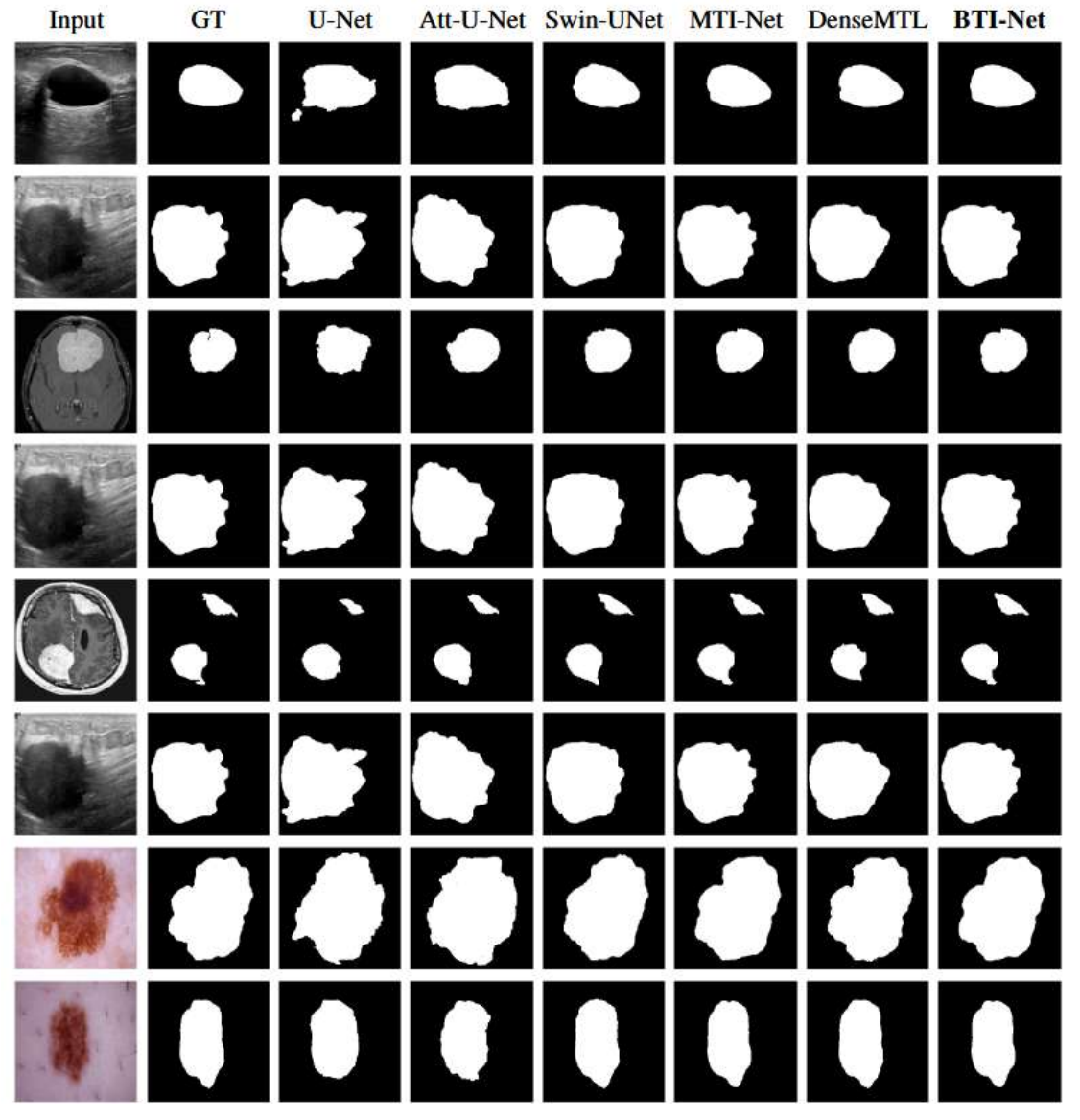}
\caption{Qualitative segmentation results.
\textbf{Rows 1--2}: BUSI (ultrasound);
\textbf{Rows 3--4}: BRISC (brain MRI);
\textbf{Rows 5--8}: HAM10000 (dermoscopy).
\textbf{Columns}: input, ground truth, U-Net, Att-U-Net, Swin-UNet,
MTI-Net, DenseMTL, BTI-Net.
Cases are selected to highlight boundary-ambiguous examples where
bidirectional cross-task interaction provides the clearest
visual improvement.}
\label{fig:qualitative}
\end{figure*}

Figure~\ref{fig:qualitative} presents representative segmentation
outputs across all three modalities.
BTI-Net recovers sharper lesion boundaries with fewer false positives,
particularly in low-contrast regions where encoder-sharing methods
produce over-segmented blobs.
Against MTI-Net and DenseMTL, the improvement is most
visible at object boundaries: the Clf$\rightarrow$Seg pathway injects
class-specific priors that suppress activations inconsistent with the
predicted lesion type, an effect absent in methods that apply fixed
per-sample blending.
Failure cases are provided in the supplementary material.

\section{Conclusion}

BTI-Net addresses a recognized gap in multi-task medical image
analysis: the absence of cross-task refinement during spatial
reconstruction.
Task Interaction Modules establish bidirectional decoder-level
communication at all four resolutions, creating a progressive closed
loop from coarse semantic exchange at $D_1$ to fine boundary
correction at $D_4$.
Uncertainty Proxy Attention gates each interaction per instance and
per level using three signals derived from feature activation
statistics, without external annotations or Bayesian overhead.

Evaluation across three diverse imaging modalities (breast ultrasound,
dermoscopy, and brain MRI) over three independent seeds demonstrates
consistent improvement over both encoder-sharing methods and prior
decoder-interaction approaches including MTI-Net and DenseMTL across
all six metric--dataset pairs.
Ablation confirms that adaptive gating contributes +2.36 IoU over fixed bidirectional interaction,
and that difficulty-aware supervision is necessary for the gate to learn a meaningful difficulty
ranking.

UPA provides competitive single-pass failure signals, matching or
exceeding MC Dropout on segmentation AUC-ROC across all benchmarks;
classification failure detection surpasses MC Dropout in AUPRC across
all three datasets, though UPA does not uniformly close the gap to
the maximum-softmax-confidence baseline, pointing toward task-specific
uncertainty heads as a direction for future work.

The framework is compatible with standard encoder backbones and
applicable to other multi-task dense prediction settings with
complementary inter-task structure.
Extending the design to 3D volumetric data and to natural-image
task pairs such as semantic segmentation with depth estimation are
natural next steps.

{\small
\bibliographystyle{ieeenat_fullname}
\bibliography{main}
}

\input{appendix_arxiv}

\end{document}

%% file: appendix_arxiv.tex
\setcounter{figure}{0}
\renewcommand{\thefigure}{A\arabic{figure}}
\setcounter{table}{0}
\setcounter{section}{0}               
\renewcommand{\thetable}{A\arabic{table}}
\renewcommand{\thesection}{\Alph{section}}

\twocolumn[{%
\begin{center}
  {\Large\bfseries BTI-Net: Bidirectional Decoder-Level Task Interaction
  via Uncertainty-Aware Gating for Multi-Task Medical Image Analysis}
  \\[8pt]
  {\large\bfseries Supplementary Material}
\end{center}
\noindent\rule{\linewidth}{0.4pt}
\vspace{4pt}
}]

\section{Theoretical Motivation for UPA Signals}
\label{app:theory}

The three signals used by UPA, cross-task alignment, segmentation
gradient energy, and classification activation spread, proxy distinct
sources of predictive uncertainty without requiring multiple stochastic
forward passes.
Each signal is justified below in terms of its relationship to model
confidence and expected error probability.

\paragraph{Cross-task alignment as epistemic uncertainty.}
When TIM produces feature updates, the residuals
$\Delta_{\mathrm{seg}} = D_\ell^{\mathrm{enh}} - D_\ell$ and
$\Delta_{\mathrm{clf}} = f_{\mathrm{clf}}^{\ell,\mathrm{enh}} -
f_{\mathrm{clf}}^{\ell}$ encode how each task's representation is modified
by the interaction.
If both residuals point in similar directions after projection, the
interaction is \emph{mutually consistent}.
Conversely, low cosine similarity signals \emph{epistemic uncertainty}:
the two branches disagree on how to incorporate cross-task information,
indicating the interaction is unreliable for this sample.
Formally, $a_\ell$ is a sample-wise estimate of directional agreement
between the two task update vectors.

\paragraph{Segmentation gradient energy as aleatoric complexity.}
$E_{\mathrm{seg}}$ measures the mean absolute first-order spatial
differences of the base decoder features.
Sharp boundaries, fine structures, and heterogeneous textures produce
large local gradients, representing \emph{aleatoric uncertainty}
inherent to the input.
This signal naturally increases at finer decoder resolutions
($D_3$, $D_4$) where boundary detail is most pronounced, making it
scale-appropriate without requiring separate scale-specific signals.

\paragraph{Classification activation spread as prediction confidence.}
$E_{\mathrm{clf}}$ captures how peaked the classification feature vector
is via its log-variance.
A high value indicates a low-entropy prediction concentrated on dominant
channels; a low value indicates a flat, uninformative embedding.
This serves as a proxy for \emph{classification confidence}, related to
the softmax margin and expected calibration error.
Extracting spread from pre-softmax features rather than class
probabilities makes it sensitive to the network's internal representation
even when final probabilities are poorly calibrated.

\paragraph{Completeness.}
Together, $a_\ell$, $E_{\mathrm{seg}}$, and $E_{\mathrm{clf}}$ span
three complementary axes: interaction consistency, scene complexity, and
prediction confidence.
These axes are non-redundant: a low-contrast image may have low gradient
energy but low classification confidence; a texture-rich image may have
high gradient energy but high inter-task alignment.
No single signal covers all three axes.

\section{Training Protocol Details}
\label{app:training_protocol}

\paragraph{Two-stage training.}
Training proceeds in two stages to ensure the UPA gate learns from a
stable main network rather than a randomly initialized predictor.

Stage~1 trains the full model end-to-end: encoder, decoder,
TIM modules, UPA modules, and both heads under the objective
$\mathcal{L} = \mathcal{L}_{\mathrm{seg}} +
\mathcal{L}_{\mathrm{clf}}$
with all parameters trainable.
Stage~1 runs for up to 50 epochs with early stopping on validation IoU
(patience~10).

Stage~2 freezes all layers \emph{except} the four UPA gate
networks and fine-tunes only the gate parameters on $\mathcal{L}_{\mathrm{gate}}$ , isolating UPA
convergence from changes in the backbone and TIM.
Stage~2 runs for up to 15 epochs with early stopping (patience~5).

\paragraph{Learning rate schedule.}
Both stages use an initial learning rate of $3\!\times\!10^{-4}$,
reduced on validation plateau by a factor of 0.5 every 5 epochs without
improvement (minimum $10^{-6}$).
Gradient norms are clipped at 1.0 to prevent instability during early
training.

\paragraph{Loss weight adaptation.}
An adaptive scheme initialises task weights at
$\lambda_{\mathrm{seg}}\!=\!0.82$, $\lambda_{\mathrm{clf}}\!=\!0.18$
and adjusts them proportionally based on relative validation loss
improvement at each epoch.
The gate regularisation weight $\lambda_{\mathrm{gate}}$ is set to~1.0
for HAM10000 and BRISC, and~0.5 for BUSI; the lower value on BUSI
reflects the smaller dataset size and higher noise in soft-IoU estimates.

\paragraph{Data splits.}
BUSI and HAM10000 are split into approximately 70/15/15 train/validation/test partitions with
stratified label distributions; BRISC follows the official test split, with the remaining training
data split 85/15 for training/validation. Three independent random seeds (42, 123, 456) are used;
each seed determines parameter initialization and augmentation order

\section{Hyperparameter Settings}
\label{app:hyper}

\begin{table}[H]
\centering
\caption{Hyperparameter settings used across experiments. Values are consistent across all three datasets unless explicitly noted.}
\label{tab:app_hyper}
\footnotesize
\begin{tabular}{p{0.4\linewidth} p{0.45\linewidth}}
\toprule
\textbf{Parameter} & \textbf{Value} \\
\midrule
\multicolumn{2}{l}{\textit{Optimisation \& Training}} \\
Optimizer & Adam \\
Initial learning rate & $3\times10^{-4}$ \\
Learning rate schedule & ReduceLROnPlateau (factor 0.5, patience 5) \\
Weight decay ($L_2$) & $10^{-5}$ \\
Max epochs (Stage 1 \& 2) & 50 / 15 \\
Early stopping patience & 10 (val IoU) / 5 (val loss) \\
Batch size (Train / Val) & 8 / 16 \\
\midrule
\multicolumn{2}{l}{\textit{Architecture}} \\
Encoder backbone & EfficientNet-B4 (ImageNet pre-trained) \\
Decoder channels & 384, 192, 96, 48 \\
Classification feature dim & 256-D (per level) \\
UPA (hidden / projection dim) & 32 / 64 \\
Dropout rate & 0.3 (classification head) \\
\midrule
\multicolumn{2}{l}{\textit{Loss Functions \& Balancing}} \\
Segmentation loss & Focal Tversky ($\gamma=0.75$) \\
Classification loss & Focal Cross-Entropy ($\gamma=2.0$) \\
Initial task weights & $\lambda_{\mathrm{seg}}=0.82$, $\lambda_{\mathrm{clf}}=0.18$ (adaptive) \\
Auxiliary loss weights & Boundary (0.25), Texture (0.15) \\
Gate weight $\lambda_{\mathrm{gate}}$ & 0.5 (BUSI); 1.0 (HAM, BRISC) \\
TIM modulation $\tau$ & 0.7 \\
\midrule
\multicolumn{2}{l}{\textit{Data Configuration}} \\
Input resolution & $224\times224$ \\
Augmentation & H/V Flips ($p=0.5$), Rotation $\pm15^\circ$ ($p=0.7$) \\
Data split & 70/15/15 (Train/Val/Test on BUSI \& HAM), patient-wise stratified (HAM); official test split, 85/15 Train/Val (BRISC) \\
\bottomrule
\end{tabular}
\end{table}

\section{Gate Training: Derivation and Justification}
\label{app:gate_deriv}

\paragraph{Difficulty-aware target derivation.}
The soft-IoU between predicted mask and ground truth is computed
differentiably:
\begin{equation}
\mathrm{IoU}_i =
  \frac{\sum_p \hat{y}_{ip}\,y_{ip}}
       {\sum_p \hat{y}_{ip} + \sum_p y_{ip}
        - \sum_p \hat{y}_{ip}\,y_{ip} + \varepsilon}.
\label{eq:soft_iou}
\end{equation}
This is min-max normalised within the mini-batch:
\begin{equation}
\tilde{\iota}_i =
  \frac{\mathrm{IoU}_i - \min_j \mathrm{IoU}_j}
       {\max_j \mathrm{IoU}_j - \min_j \mathrm{IoU}_j + \varepsilon},
\label{eq:iou_norm}
\end{equation}
and inverted to assign target~1 to the hardest sample:
\begin{equation}
t_i^{\mathrm{seg}} = 1 - \tilde{\iota}_i.
\label{eq:target_full}
\end{equation}
For classification, the maximum predicted class probability
$c_i = \max_k p_{ik}$ is analogously normalised and inverted:
\begin{equation}
t_i^{\mathrm{clf}} =
  1 - \frac{c_i - \min_j c_j}{\max_j c_j - \min_j c_j + \varepsilon}.
\label{eq:clf_target}
\end{equation}

\paragraph{Gate regularization loss.}
The gate is trained by minimizing BCE between each target and its
gate weight, summed over both tasks and averaged over all four UPA
levels:
\begin{equation}
\mathcal{L}_{\mathrm{gate}} =
  \frac{\lambda_{\mathrm{gate}}}{4}
  \sum_{\ell=1}^{4}\!
  \bigl(
  \mathrm{BCE}(t^{\mathrm{seg}},w_{\mathrm{seg}}^\ell)
  +\mathrm{BCE}(t^{\mathrm{clf}},w_{\mathrm{clf}}^\ell)
  \bigr).
\label{eq:lgate}
\end{equation}

\paragraph{Convergence analysis.}
The BCE loss has a global minimum at $w_i = t_i$.
Since $t_i^{\mathrm{seg}}$ is computed from $\mathrm{sg}[\hat{y}]$
(stop-gradient), the gate loss does not back-propagate through the
segmentation predictions, preventing the gate from gaming the
soft-IoU score rather than tracking genuine sample difficulty.
The two-stage protocol further ensures the gate is fine-tuned after
the main network has converged, so targets are computed from a stable
predictor.

\section{UPA Uncertainty Analysis}
\label{app:upa}

This section provides the complete per-dataset signal-quality tables and
figures for the Uncertainty Proxy Attention (UPA) analysis.
For BUSI, the single-level ROC curves, precision-recall curves, and TIM
enhancement figure appear in the main paper; the remaining analysis is
reported here.
For BRISC and HAM10000 all figures are included.
All metrics are computed on the test set used for main-paper
segmentation and classification evaluation.

\paragraph{MC Dropout configuration.}
Dropout (rate~0.3, classification head only) is applied at inference;
uncertainty estimates average 30 stochastic forward passes without
post-hoc calibration.

\subsection{Cross-Dataset Summary}
\label{app:upa_summary}

\begin{figure}[h]
\centering
\includegraphics[width=0.90\linewidth]{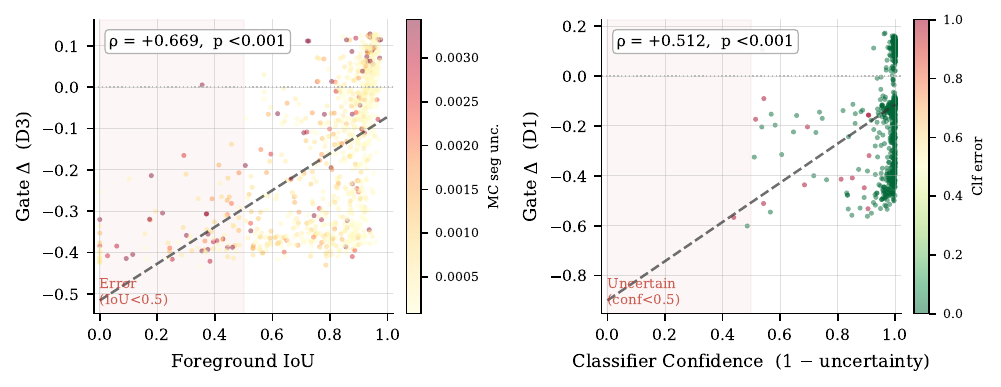}
\caption{BRISC Gate $\delta$ interpretability.
\emph{Left}: $\rho({\delta},\text{IoU}) = +0.669$ at D3 ($p<0.001$).
\emph{Right}: $\rho({\delta},\text{conf}) = +0.512$ at D1 ($p<0.001$).}
\label{fig:brisc_interp}
\end{figure}

\begin{figure}[h]
\centering
\includegraphics[width=0.90\linewidth]{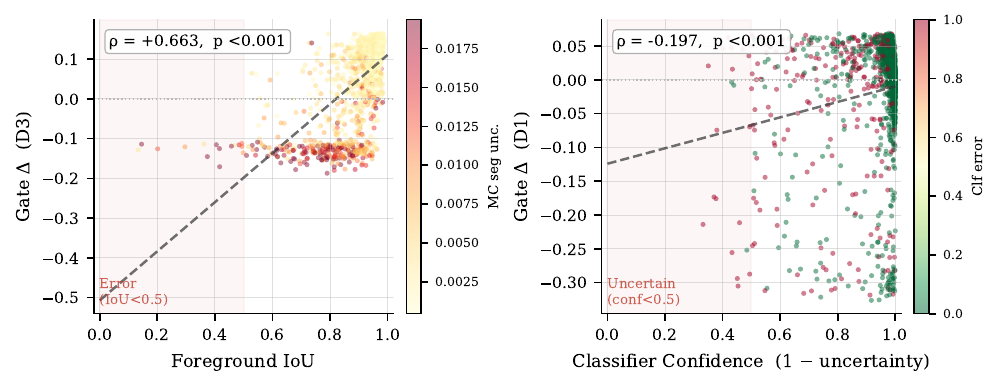}
\caption{HAM Gate $\delta$ interpretability.
\emph{Left}: $\rho({\delta},\text{IoU}) = +0.663$ at D3 ($p<0.001$),
consistent with BUSI and BRISC.
\emph{Right}: $\rho({\delta},\text{conf}) = -0.197$ at D1 ($p<0.001$);
significant but weaker than seg panel, reflecting limited clf
discriminability at this class-imbalance level.}
\label{fig:ham_interp}
\end{figure}

Table~\ref{tab:app_upa_summary} summarises segmentation-failure detection
across all three datasets using the AUC-weighted combined gate signal and
the best single-level signal per dataset.
Best-level results report the strongest signal at the strongest decoder
level; since the same level ranks highest consistently across all three
random seeds, these results characterize a structural property of the
architecture and are not sensitive to the specific test partition.
The AUC-weighted combination assigns weights from per-level AUC-ROC
scores and should be interpreted as a diagnostic summary of signal
quality rather than a tuned deployment protocol. We further define the gate differential
$\delta_\ell = w_{\mathrm{clf}} - w_{\mathrm{seg}}$,
which quantifies the relative task preference of the gate at level $\ell$.
Negative values indicate the gate allocates more enhancement budget to
segmentation, empirically correlating with hard segmentation samples
(Figs.~\ref{fig:busi_interp}~\ref{fig:ham_interp}~\ref{fig:brisc_interp}).

\begin{table}[t]
\centering
\caption{Segmentation-failure detection AUC-ROC and AUPRC.
\textit{UPA combined}: AUC-weighted fusion of $-\delta$ across
$D_1$--$D_4$. 
\textit{UPA best}: strongest signal at the strongest decoder level
per dataset.
Random AUPRC equals the per-dataset segmentation-error rate.}
\label{tab:app_upa_summary}
\footnotesize
\setlength{\tabcolsep}{2pt}
\begin{tabular}{@{} l c c c c c c @{}}
\toprule
 & \multicolumn{2}{c}{\textbf{MC (30$\times$)}}
 & \multicolumn{2}{c}{\textbf{UPA combined}}
 & \multicolumn{2}{c}{\textbf{UPA best level}} \\
\cmidrule(lr){2-3}\cmidrule(lr){4-5}\cmidrule(lr){6-7}
\textbf{Dataset} & AUC & AP & AUC & AP & AUC (signal, level) & AP \\
\midrule
BUSI   & 0.751 & 0.431 & \textbf{0.807} & \textbf{0.610}
        & \textbf{0.824} ($w_\text{seg}$, $D_2$) & \textbf{0.602} \\
BRISC  & 0.702 & 0.238 & \textbf{0.806} & \textbf{0.332}
        & \textbf{0.852} ($-\delta$, $D_3$)     & \textbf{0.470} \\
HAM10000 & 0.891 & 0.151 & \textbf{0.927} & 0.073
          & \textbf{0.934} ($w_\text{seg}$, $D_3$) & 0.079 \\
\bottomrule
\multicolumn{7}{@{}p{\columnwidth}@{}}{\footnotesize HAM10000 AUPRC is uninformative: seg-error rate $=$ 0.010; random baseline $\approx$ 0.010. All HAM10000 uncertainty claims therefore rest on AUC-ROC only.}
\end{tabular}
\end{table}

\subsection{ BUSI - Breast Ultrasound}
\label{app:upa_busi}

\paragraph{Signal quality.}
Table~\ref{tab:busi_sig_seg} reports segmentation-error detection
AUC-ROC across all four decoder levels and all UPA signals.
Table~\ref{tab:busi_sig_clf} reports classification-error detection.
Entries marked $^*$ beat the MC Dropout reference.

\begin{table}[h]
\centering
\caption{BUSI Seg-error AUC-ROC.  MC ref = 0.751.
$^*$ = beats MC.  Bold = best per level.}
\label{tab:busi_sig_seg}
\footnotesize
\setlength{\tabcolsep}{4pt}
\begin{tabular}{lcccc}
\toprule
\textbf{Signal} & \textbf{D1 (14$\times$14)} & \textbf{D2 (28$\times$28)}
               & \textbf{D3 (56$\times$56)} & \textbf{D4 (112$\times$112)} \\
\midrule
$-\delta$      & 0.674 & $\mathbf{0.797}^*$ & 0.461 & 0.687 \\
$-$align       & 0.236 & $0.803^*$          & 0.728 & 0.206 \\
seg\_energy    & 0.331 & 0.267              & 0.285 & 0.330 \\
$w_\text{seg}$ & $0.767^*$ & $\mathbf{0.824}^*$ & 0.682 & $0.785^*$ \\
\midrule
MC Dropout     & \multicolumn{4}{c}{0.751 (30 passes)} \\
\bottomrule
\end{tabular}
\end{table}

\begin{table}[h]
\centering
\caption{BUSI  Clf-error AUC-ROC.  MC ref = 0.728; 1-conf = 0.834.
$^*$ = beats MC.}
\label{tab:busi_sig_clf}
\footnotesize
\setlength{\tabcolsep}{4pt}
\begin{tabular}{lcccc}
\toprule
\textbf{Signal} & \textbf{D1} & \textbf{D2} & \textbf{D3} & \textbf{D4} \\
\midrule
$-\delta$      & 0.435 & 0.523 & 0.306 & 0.580 \\
clf\_energy    & 0.286 & 0.223 & 0.244 & 0.351 \\
$w_\text{clf}$ & $0.768^*$ & 0.716 & 0.725 & $\mathbf{0.850}^*$ \\
$-$align       & 0.402 & $0.754^*$ & 0.511 & 0.201 \\
\midrule
MC Dropout     & \multicolumn{4}{c}{0.728} \\
1-conf         & \multicolumn{4}{c}{0.834} \\
\bottomrule
\end{tabular}
\end{table}

\noindent\textbf{Combined gate output.}
Table~\ref{tab:busi_combined} shows the combined gate output
and MC Dropout for segmentation failure detection.

\begin{table}[h]
\centering
\caption{BUSI  Combined UPA gate vs MC Dropout, seg error.}
\label{tab:busi_combined}
\footnotesize
\setlength{\tabcolsep}{4pt}
\begin{tabular}{lccc}
\toprule
\textbf{Method} & \textbf{AUC} & \textbf{AUPRC} & \textbf{ECE} \\
\midrule
MC Dropout (ref.)         & 0.751 & 0.431 & 0.102 \\
UPA $-\delta$ mean D1--D4 & 0.772 & 0.554 & 0.363 \\
UPA $-\delta$ AUC-weighted& \textbf{0.807} & \textbf{0.610} & 0.237 \\
UPA $-\delta$ best D2     & 0.797 & 0.573 & 0.172 \\
\bottomrule
\end{tabular}
\end{table}

\noindent\textbf{Gate interpretability.}
Table~\ref{tab:busi_spearman} reports Spearman $\rho$ between UPA signals
and MC Dropout uncertainty. $\checkmark$~=~$|\rho|>0.3$, $p<0.05$.

\begin{table}[h]
\centering
\caption{BUSI  Spearman $\rho$ vs MC Dropout.}
\label{tab:busi_spearman}
\footnotesize
\setlength{\tabcolsep}{3pt}
\begin{tabular}{lcccc}
\toprule
\textbf{Pair} & \textbf{D1} & \textbf{D2} & \textbf{D3} & \textbf{D4} \\
\midrule
seg\_energy vs MC seg
  & $\checkmark$$-$0.586 & $\checkmark$$-$0.653
  & $\checkmark$$-$0.689 & $\checkmark$$-$0.661 \\
$\delta$ vs MC seg
  & $+$0.203 & $-$0.055 & $\checkmark$$+$0.601 & $\checkmark$$+$0.435 \\
align vs MC MI
  & $-$0.014 & $-$0.269 & $+$0.097 & $+$0.154 \\
$\delta$ vs MC ent
  & $+$0.262 & $+$0.091 & $\checkmark$$+$0.334 & $+$0.147 \\
\bottomrule
\multicolumn{5}{l}{\footnotesize Note: seg\_energy is anti-correlated with
  MC seg on BUSI (negative $\rho$),}\\
\multicolumn{5}{l}{\footnotesize contrast to BRISC. See \ref{app:upa_disc}.}
\end{tabular}
\end{table}

\paragraph{Gate $\delta$ wrong vs correct.}
D1: $t=-0.52$, $p=0.602$; D4: $t=-0.55$, $p=0.585$.
Neither is significant; this result is not claimed as supportive evidence.

\begin{figure}[h]
\centering
\includegraphics[width=0.90\linewidth]{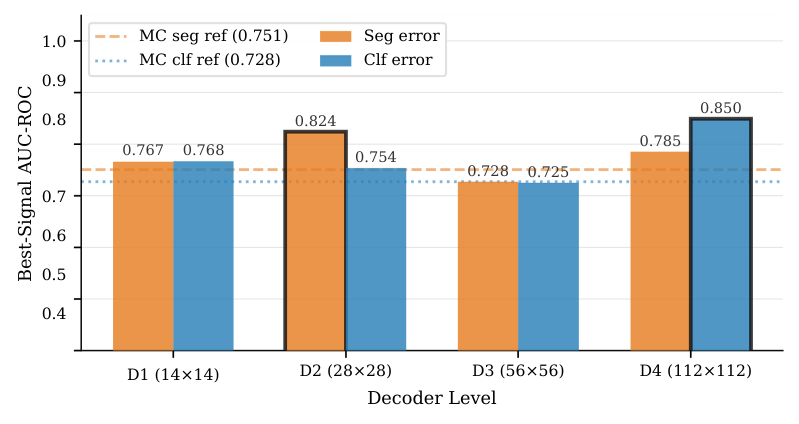}
\caption{BUSI  Scale sensitivity.  Each bar shows the best-signal
AUC-ROC at that decoder level.  Bold border = best level.
Dashed/dotted lines = MC Dropout reference.}
\label{fig:busi_scale}
\end{figure}

\begin{figure}[h]
\centering
\includegraphics[width=0.90\linewidth]{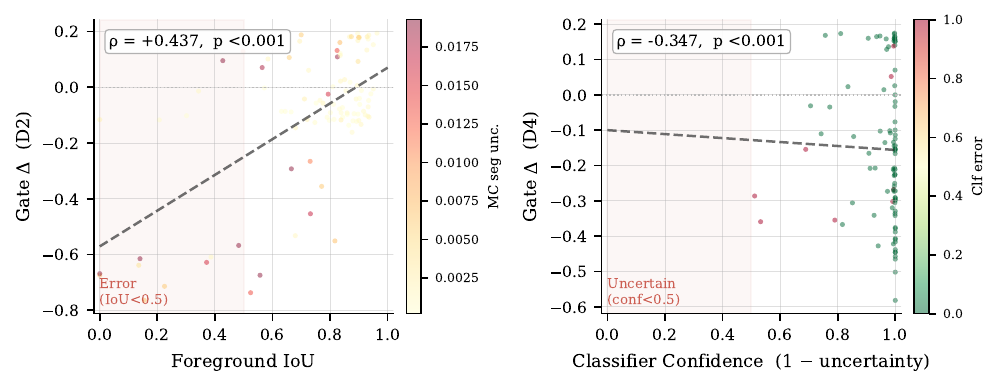}
\caption{BUSI  Gate $\delta$ interpretability.
\emph{Left}: $\delta$ vs foreground IoU at D2
($\rho=+0.437$, $p<0.001$); colour = MC seg uncertainty.
\emph{Right}: $\delta$ vs classifier confidence at D4
($\rho=-0.347$, $p<0.001$); colour = clf error.}
\label{fig:busi_interp}
\end{figure}

\begin{figure}[h]
\centering
\includegraphics[width=0.90\linewidth]{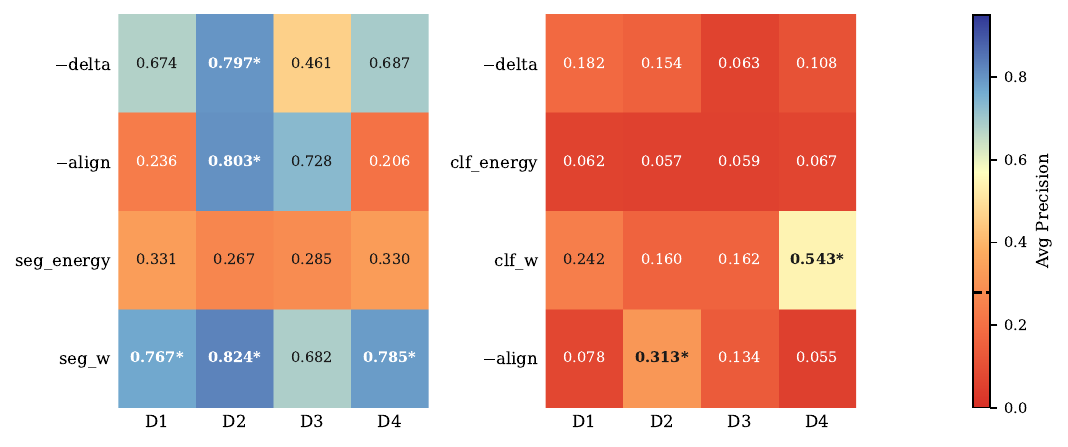}
\caption{BUSI  Signal-quality heatmap.
\emph{Left}: seg-error AUC-ROC per signal $\times$ level.
\emph{Right}: clf-error Avg Precision.
$*$ = beats MC Dropout; dashed line on colorbar = MC reference.}
\label{fig:busi_heatmap}
\end{figure}

\begin{figure}[h]
\centering
\includegraphics[width=0.55\linewidth]{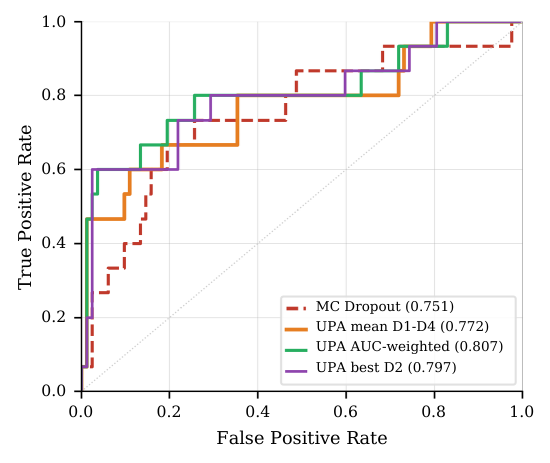}
\caption{BUSI  Combined gate ROC.
AUC-weighted fusion (green) achieves 0.807 vs MC Dropout 0.751.}
\label{fig:busi_combined_roc}
\end{figure}

Figure~\ref{fig:busi_scale} shows scale sensitivity across decoder levels.
Figure~\ref{fig:busi_interp} shows gate~$\delta$ versus foreground IoU and
classifier confidence.
Figure~\ref{fig:busi_heatmap} shows the full signal-quality heatmap.
Figure~\ref{fig:busi_combined_roc} shows the combined gate ROC.

\subsection{BRISC - Brain MRI}
\label{app:upa_brisc}

\begin{table}[H]
\centering
\caption{BRISC Seg-error AUC-ROC.  MC ref = 0.702.
$^*$ = beats MC.}
\label{tab:brisc_sig_seg}
\footnotesize
\setlength{\tabcolsep}{4pt}
\begin{tabular}{lcccc}
\toprule
\textbf{Signal} & \textbf{D1} & \textbf{D2} & \textbf{D3} & \textbf{D4} \\
\midrule
$-\delta$      & 0.413 & 0.556 & $\mathbf{0.852}^*$ & 0.515 \\
$-$align       & 0.579 & 0.512 & $0.823^*$          & 0.467 \\
seg\_energy    & 0.603 & 0.566 & 0.509              & 0.445 \\
$w_\text{seg}$ & 0.282 & 0.482 & $0.834^*$          & 0.518 \\
\midrule
MC Dropout     & \multicolumn{4}{c}{0.702} \\
\bottomrule
\end{tabular}
\end{table}

\begin{table}[h]
\centering
\caption{BRISC Clf-error AUC-ROC.
MC ref = 0.397 (anti-correlated entropy; 1-conf = 0.920).
$^*$ = beats MC.}
\label{tab:brisc_sig_clf}
\footnotesize
\setlength{\tabcolsep}{4pt}
\begin{tabular}{lcccc}
\toprule
\textbf{Signal} & \textbf{D1} & \textbf{D2} & \textbf{D3} & \textbf{D4} \\
\midrule
$-\delta$      & $\mathbf{0.753}^*$ & $0.600^*$ & $0.597^*$ & $0.683^*$ \\
clf\_energy    & $0.516^*$ & $0.509^*$ & $0.518^*$ & $0.426^*$ \\
$w_\text{clf}$ & $0.509^*$ & $0.537^*$ & $0.575^*$ & 0.395 \\
$-$align       & 0.306     & 0.393     & $0.607^*$ & 0.279 \\
\midrule
MC Dropout     & \multicolumn{4}{c}{0.397 (miscalibrated; see note)} \\
1-conf         & \multicolumn{4}{c}{0.920} \\
\bottomrule
\multicolumn{5}{l}{\footnotesize MC entropy anti-correlated on BRISC;
  all UPA signals beat MC trivially.}\\
\multicolumn{5}{l}{\footnotesize Do not interpret as UPA clf victory
  over a valid baseline.}
\end{tabular}
\end{table}

\begin{table}[h]
\centering
\caption{BRISC Combined UPA gate vs MC Dropout, seg error.}
\label{tab:brisc_combined}
\footnotesize
\setlength{\tabcolsep}{4pt}
\begin{tabular}{lccc}
\toprule
\textbf{Method} & \textbf{AUC} & \textbf{AUPRC} & \textbf{ECE} \\
\midrule
MC Dropout (ref.)         & 0.702 & 0.238 & 0.091 \\
UPA $-\delta$ mean D1--D4 & 0.635 & 0.198 & 0.281 \\
UPA $-\delta$ AUC-weighted& \textbf{0.806} & \textbf{0.332} & 0.395 \\
UPA $-\delta$ best D3     & 0.852 & 0.470 & 0.420 \\
\bottomrule
\end{tabular}
\end{table}

\begin{table}[h]
\centering
\caption{BRISC Gate $\delta$ wrong vs correct clf.
Both levels significant.}
\label{tab:brisc_wrong_correct}
\footnotesize
\setlength{\tabcolsep}{3pt}
\begin{tabular}{lcccccc}
\toprule
 & $n_\text{wrong}$ & $\bar\delta_\text{wrong}$
 & $n_\text{correct}$ & $\bar\delta_\text{correct}$
 & $t$ & $p$ \\
\midrule
D1 & 14 & $-$0.309 & 846 & $-$0.129 & $-$3.29 & 0.001 $\checkmark$ \\
D4 & 14 & $-$0.139 & 846 & $-$0.036 & $-$2.48 & 0.014 $\checkmark$ \\
\bottomrule
\end{tabular}
\end{table}

\begin{table}[h]
\centering
\caption{BRISC Spearman $\rho$ vs MC Dropout.
$\checkmark$~=~$|\rho|>0.3$, $p<0.05$.}
\label{tab:brisc_spearman}
\footnotesize
\setlength{\tabcolsep}{3pt}
\begin{tabular}{lcccc}
\toprule
\textbf{Pair} & \textbf{D1} & \textbf{D2} & \textbf{D3} & \textbf{D4} \\
\midrule
seg\_energy vs MC seg
  & $+$0.252 & $+$0.212 & $+$0.258 & $-$0.094 \\
$\delta$ vs MC seg
  & $-$0.114 & $+$0.055 & $-$0.013 & $\checkmark$$+$0.370 \\
align vs MC MI
  & $+$0.116 & $+$0.186 & $-$0.090 & $+$0.085 \\
$\delta$ vs MC ent
  & $-$0.210 & $+$0.031 & $+$0.029 & $+$0.160 \\
\bottomrule
\end{tabular}
\end{table}

\begin{figure}[h]
\centering
\includegraphics[width=\linewidth]{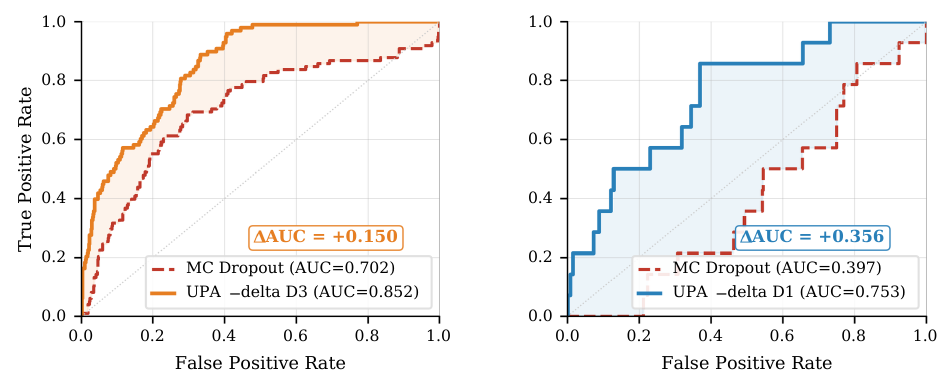}
\caption{BRISC Single-level ROC.
\emph{Left}: seg-error, UPA $-\delta$ D3 (AUC=0.852) vs MC (0.702),
$\Delta$AUC = +0.150.
\emph{Right}: clf-error, UPA $-\delta$ D1 (AUC=0.753) vs MC (0.397);
note MC entropy is anti-correlated on BRISC.}
\label{fig:brisc_roc}
\end{figure}

\begin{figure}[h]
\centering
\includegraphics[width=\linewidth]{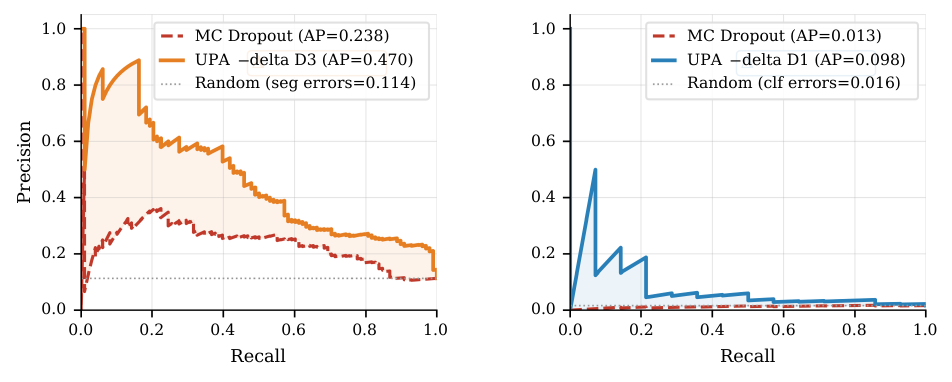}
\caption{BRISC Precision-recall curves.
\emph{Left}: seg-error, UPA AP=0.470 vs MC AP=0.238.
\emph{Right}: clf-error shown for completeness; MC baseline is
unreliable (random clf-error rate = 0.016).}
\label{fig:brisc_pr}
\end{figure}

\begin{figure}[h]
\centering
\includegraphics[width=0.90\linewidth]{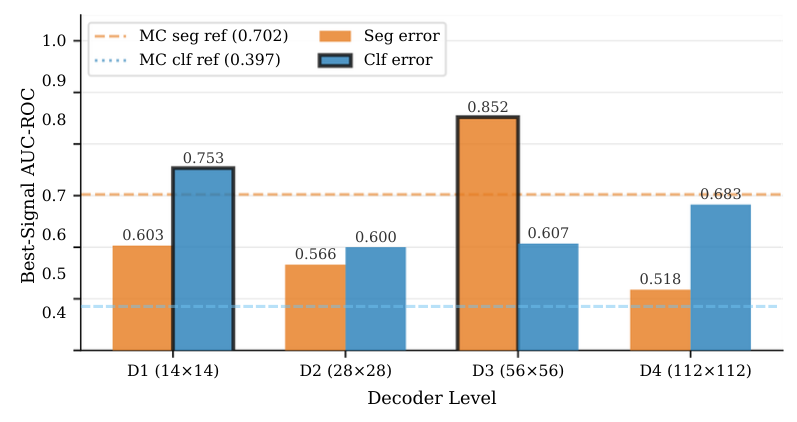}
\caption{BRISC Scale sensitivity.  D3 is the best level for seg.
Clf bars all exceed the miscalibrated MC reference (0.397).}
\label{fig:brisc_scale}
\end{figure}

\begin{figure}[h]
\centering
\includegraphics[width=0.90\linewidth]{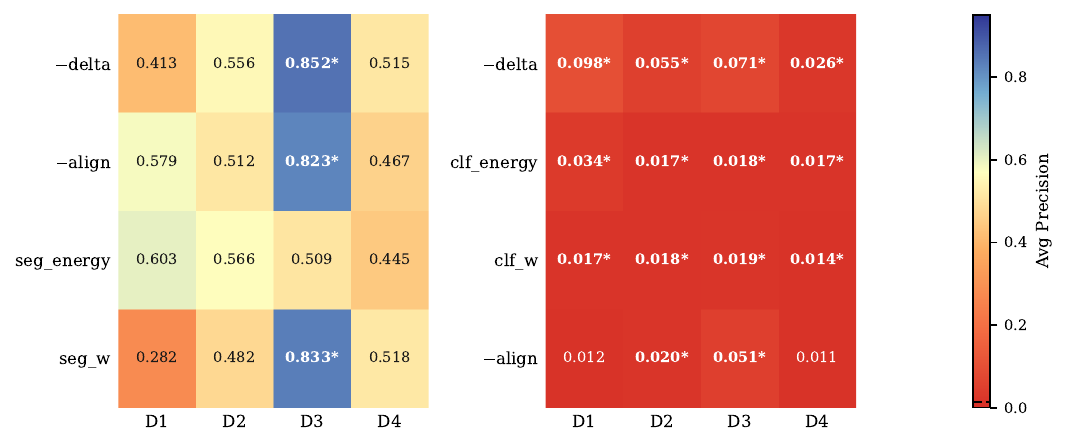}
\caption{BRISC Signal-quality heatmap.
Seg (left): D3 is strongly informative; D1/D2/D4 are near chance.
Clf (right): all signals near random baseline given tiny clf error
rate ($\approx$1.6\%).}
\label{fig:brisc_heatmap}
\end{figure}

\begin{figure}[h]
\centering
\includegraphics[width=0.55\linewidth]{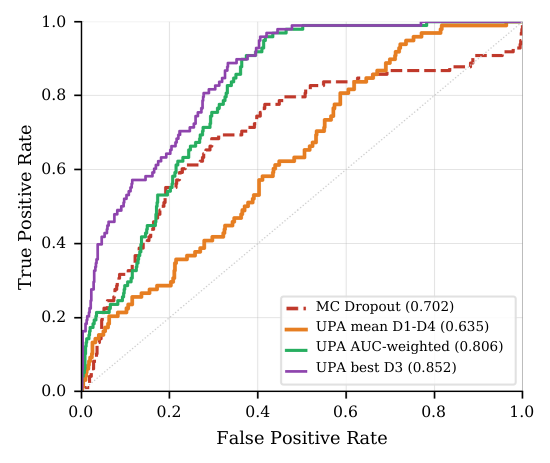}
\caption{BRISC Combined gate ROC.
AUC-weighted fusion achieves 0.806; simple mean (0.635) falls below
MC because D1/D2/D4 are below chance and drag the average down.
This motivates the AUC-weighted combination strategy.}
\label{fig:brisc_combined_roc}
\end{figure}

\begin{figure}[h]
\centering
\includegraphics[width=0.90\linewidth]{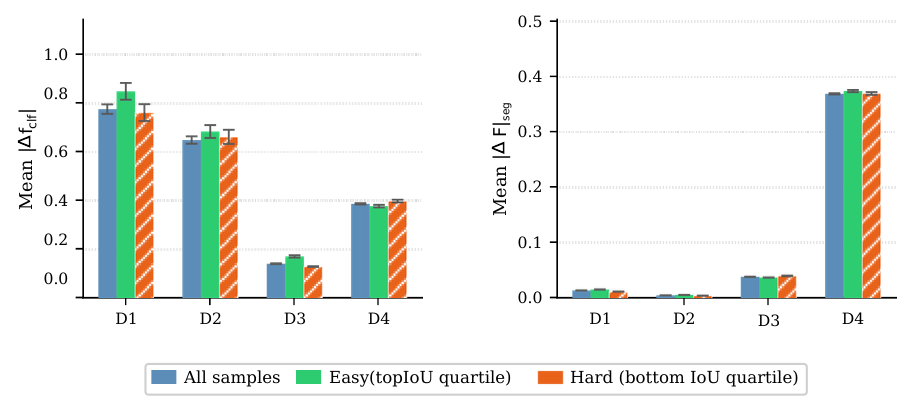}
\caption{BRISC TIM enhancement magnitudes.
\emph{Left} (Seg$\to$Clf): hard samples show comparable or larger
enhancement than easy samples at D1--D2 but converge at D4.
\emph{Right} (Clf$\to$Seg): magnitudes are small at D1--D2, growing
at D3--D4; hard/easy difference is negligible.}
\label{fig:brisc_tim}
\end{figure}

\subsection{ HAM10000 - Dermoscopy}
\label{app:upa_ham}
This subsection presents the comprehensive failure detection tables, scale-sensitivity charts, and gate interpretability figures for the HAM10000 dermoscopy dataset.

\begin{table}[h]
\centering
\caption{HAM Seg-error AUC-ROC.  MC ref = 0.891.
$^*$ = beats MC.}
\label{tab:ham_sig_seg}
\footnotesize
\setlength{\tabcolsep}{4pt}
\begin{tabular}{lcccc}
\toprule
\textbf{Signal} & \textbf{D1} & \textbf{D2} & \textbf{D3} & \textbf{D4} \\
\midrule
$-\delta$      & 0.814 & $0.916^*$ & 0.885 & $0.922^*$ \\
$-$align       & $0.898^*$ & 0.081 & 0.062 & $0.924^*$ \\
seg\_energy    & 0.333 & 0.276 & 0.143 & 0.415 \\
$w_\text{seg}$ & $0.901^*$ & $0.918^*$ & $\mathbf{0.934}^*$ & $0.926^*$ \\
\midrule
MC Dropout     & \multicolumn{4}{c}{0.891} \\
\bottomrule
\end{tabular}
\end{table}

\begin{table}[h]
\centering
\caption{HAM Clf-error AUC-ROC.  MC ref = 0.575;
1-conf = 0.834.  $^*$ = beats MC.}
\label{tab:ham_sig_clf}
\footnotesize
\setlength{\tabcolsep}{4pt}
\begin{tabular}{lcccc}
\toprule
\textbf{Signal} & \textbf{D1} & \textbf{D2} & \textbf{D3} & \textbf{D4} \\
\midrule
$-\delta$      & 0.435 & $0.597^*$ & $0.631^*$ & $0.736^*$ \\
clf\_energy    & 0.260 & $0.589^*$ & 0.514     & 0.302 \\
$w_\text{clf}$ & $\mathbf{0.766}^*$ & 0.513 & $0.609^*$ & 0.518 \\
$-$align       & $0.707^*$ & 0.413 & 0.376 & $0.640^*$ \\
\midrule
MC Dropout     & \multicolumn{4}{c}{0.575} \\
1-conf         & \multicolumn{4}{c}{0.834} \\
\bottomrule
\end{tabular}
\end{table}

\begin{table}[h]
\centering
\caption{HAM Combined UPA gate vs MC Dropout, seg error.
Note: HAM AUPRC is unreliable (seg error rate = 1.0\%).}
\label{tab:ham_combined}
\footnotesize
\setlength{\tabcolsep}{4pt}
\begin{tabular}{lccc}
\toprule
\textbf{Method} & \textbf{AUC} & \textbf{AUPRC$^\dagger$} & \textbf{ECE} \\
\midrule
MC Dropout (ref.)         & 0.891 & 0.151 & 0.072 \\
UPA $-\delta$ mean D1--D4 & 0.925 & 0.071 & 0.214 \\
UPA $-\delta$ AUC-weighted& \textbf{0.927} & 0.073 & 0.212 \\
UPA $-\delta$ best D4     & 0.922 & 0.080 & 0.282 \\
\bottomrule
\multicolumn{4}{l}{\footnotesize $^\dagger$ AUPRC not meaningful:
  random baseline $\approx$ 0.010 = error rate.}
\end{tabular}
\end{table}

\begin{table}[h]
\centering
\caption{HAM Gate $\delta$ wrong vs correct clf.
D4 is strongly significant; D1 is not.}
\label{tab:ham_wrong_correct}
\footnotesize
\setlength{\tabcolsep}{3pt}
\begin{tabular}{lcccccc}
\toprule
 & $n_\text{wrong}$ & $\bar\delta_\text{wrong}$
 & $n_\text{correct}$ & $\bar\delta_\text{correct}$
 & $t$ & $p$ \\
\midrule
D1 & 331 & $-$0.025 & 1160 & $-$0.016 & $-$1.75 & 0.080 \\
D4 & 331 & $+$0.223 & 1160 & $+$0.259 & $-$9.13
   & $<$0.001 $\checkmark$ \\
\bottomrule
\multicolumn{7}{l}{\footnotesize At D4, wrong-class samples carry
  \emph{lower} $\delta$ than correct ones,}\\
\multicolumn{7}{l}{\footnotesize consistent with greater seg--clf
  task tension on hard samples.}
\end{tabular}
\end{table}

\begin{table}[h]
\centering
\caption{HAM Spearman $\rho$ vs MC Dropout.
$\checkmark$~=~$|\rho|>0.3$, $p<0.05$.
Note mixed-sign correlations at D2--D3 reflect inverted signal
regime at near-zero error rates.}
\label{tab:ham_spearman}
\footnotesize
\setlength{\tabcolsep}{3pt}
\begin{tabular}{lcccc}
\toprule
\textbf{Pair} & \textbf{D1} & \textbf{D2} & \textbf{D3} & \textbf{D4} \\
\midrule
seg\_energy vs MC seg
  & $\checkmark$$+$0.388 & $-$0.112 & $\checkmark$$-$0.440 & $-$0.253 \\
$\delta$ vs MC seg
  & $\checkmark$$-$0.387 & $\checkmark$$-$0.663 & $\checkmark$$-$0.540
  & $-$0.103 \\
align vs MC MI
  & $\checkmark$$-$0.522 & $+$0.287 & $\checkmark$$+$0.323
  & $\checkmark$$-$0.387 \\
$\delta$ vs MC ent
  & $-$0.141 & $\checkmark$$-$0.312 & $\checkmark$$-$0.313 & $-$0.252 \\
\bottomrule
\end{tabular}
\end{table}

\begin{figure}[h]
\centering
\includegraphics[width=\linewidth]{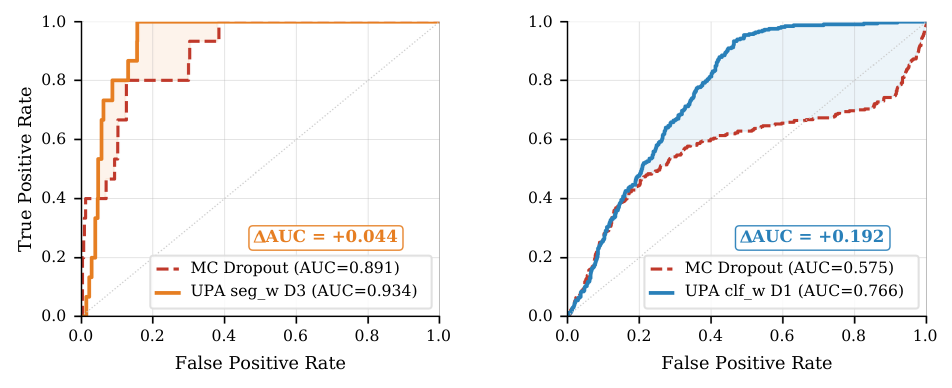}
\caption{HAM Single-level ROC.
\emph{Left}: seg-error, $w_\text{seg}$ D3 (AUC=0.934) vs MC (0.891),
$\Delta$AUC = +0.044.
\emph{Right}: clf-error, $w_\text{clf}$ D1 (AUC=0.766) vs MC (0.575),
$\Delta$AUC = +0.192.}
\label{fig:ham_roc}
\end{figure}

\begin{figure}[h]
\centering
\includegraphics[width=\linewidth]{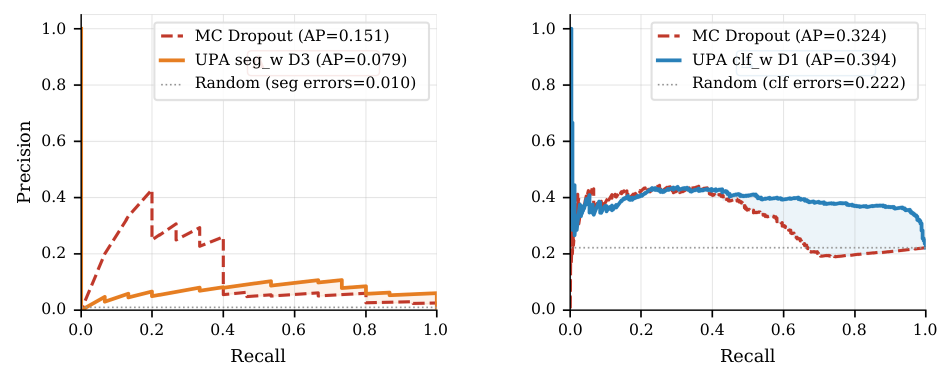}
\caption{HAM Precision-recall curves.
\emph{Left}: seg-error PR; UPA AP = 0.079 vs MC AP = 0.151;
both near random (0.010) AUPRC is not a meaningful metric
at 1\% error prevalence.
\emph{Right}: clf-error PR; UPA $w_\text{clf}$ D1 AP = 0.394
vs MC AP = 0.324.}
\label{fig:ham_pr}
\end{figure}

\begin{figure}[h]
\centering
\includegraphics[width=0.90\linewidth]{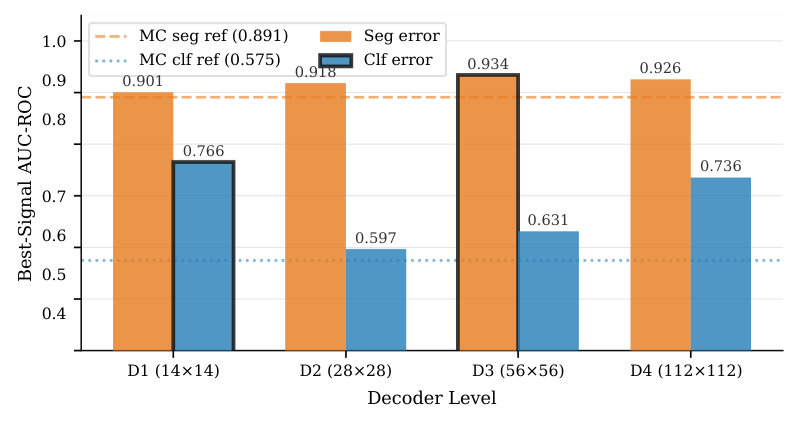}
\caption{HAM Scale sensitivity.
$w_\text{seg}$ consistently exceeds MC reference (0.891) at all
four decoder levels.  Clf signal ($w_\text{clf}$) peaks at D1.}
\label{fig:ham_scale}
\end{figure}

\begin{figure}[h]
\centering
\includegraphics[width=0.90\linewidth]{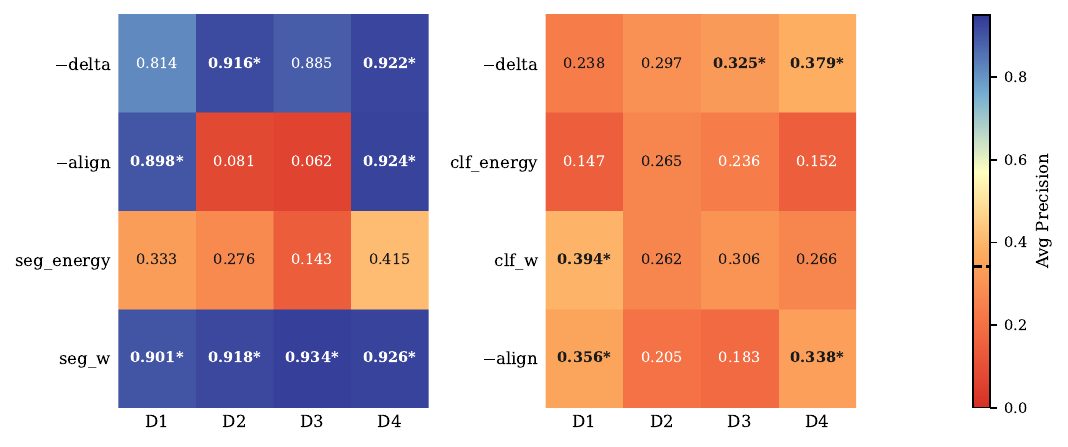}
\caption{HAM Signal-quality heatmap.
$w_\text{seg}$ (bottom row, left panel) shows strong, consistent
signal across all decoder levels, the only dataset where D4 also
beats MC comfortably.}
\label{fig:ham_heatmap}
\end{figure}

\begin{figure}[h]
\centering
\includegraphics[width=0.55\linewidth]{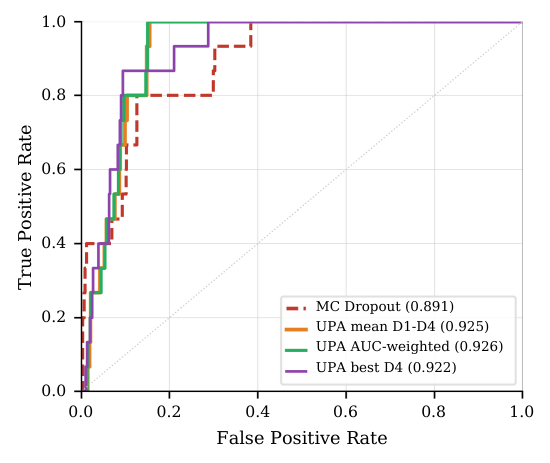}
\caption{HAM Combined gate ROC.
Both mean and AUC-weighted combinations exceed MC (0.891).
The simple mean performs comparably to the weighted version,
reflecting that $-\delta$ is informative at all four levels on HAM.}
\label{fig:ham_combined_roc}
\end{figure}

\begin{figure}[h]
\centering
\includegraphics[width=0.90\linewidth]{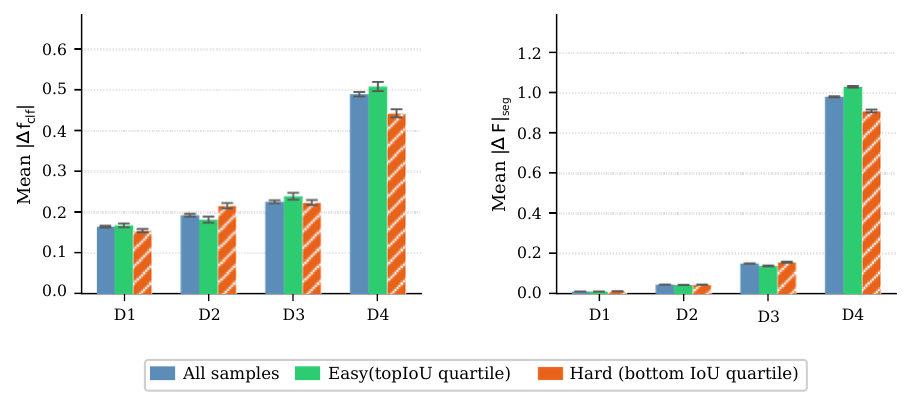}
\caption{HAM TIM enhancement magnitudes.
\emph{Left} (Seg$\to$Clf): hard samples (orange) receive lower
Seg$\to$Clf enhancement than easy samples at all levels, unlike
BUSI and BRISC, suggesting TIM adaptive routing is modality
dependent.
\emph{Right} (Clf$\to$Seg): easy samples receive marginally larger
enhancement; hard/easy gap is small.}
\label{fig:ham_tim}
\end{figure}

\subsection{Discussion of Cross-Dataset Anomalies}
\label{app:upa_disc}

\paragraph{Negative Spearman correlations on BUSI and HAM.}
On BUSI, seg\_energy is negatively correlated with MC seg uncertainty
across all decoder levels (Table~\ref{tab:busi_spearman}).
On HAM, $\delta$ and seg\_energy show mixed-sign correlations at
D2--D3 (Table~\ref{tab:ham_spearman}).
Both anomalies occur on datasets where the model achieves high
predictive accuracy (BUSI Clf Acc~=~93.16\%; HAM Clf Acc~=~87.8\%).
Thus in high-accuracy regimes, MC Dropout variance
is dominated by the network's weight uncertainty on easy samples
(which are numerous), while UPA gate signals are tuned toward
hard-sample difficulty. The resulting inversion in rank-order
correlation does not contradict the main detection-AUC results:
$w_\text{seg}$ still correctly ranks hard segmentation samples above
easy ones (positive Spearman $\rho$ with IoU, all datasets).

\paragraph{HAM AUPRC.}
The segmentation error rate on HAM is approximately 1\% (random
baseline AUPRC $\approx$ 0.010). AUPRC at this prevalence is
dominated by chance precision at high recall thresholds and is
not a reliable metric. All HAM uncertainty claims are therefore
based exclusively on AUC-ROC.

\paragraph{BRISC MC Dropout clf failure.}
MC Dropout entropy is anti-correlated with classification errors
on BRISC (AUC~=~0.397, below chance). This reflects a known failure
mode of entropy-based MC Dropout when the model is overconfident
on the wrong class. The 1-confidence baseline (AUC~=~0.920) succeeds
because the model's softmax output is well-calibrated despite the
entropy failure. UPA clf signals ($-\delta$ at D1, AUC~=~0.753)
partially recover clf uncertainty detection without requiring
post-hoc calibration, but do not match 1-conf on this dataset.

\section{Per-Dataset Segmentation Metrics}
\label{app:metrics}

Table~\ref{tab:app_seg_summary} reports segmentation performance on
positive-mask test cases for all three datasets.
Metrics are computed per-image and averaged; BUSI excludes the Normal
class (empty masks) from this computation.
HAM10000 reports over all test cases as every image contains a lesion.

\begin{table}[h]
\centering
\caption{Segmentation on positive-mask test cases.}
\label{tab:app_seg_summary}
\footnotesize
\setlength{\tabcolsep}{4pt}
\begin{tabular}{lccccc}
\toprule
\textbf{Dataset} & \textbf{Dice} & \textbf{IoU}
  & \textbf{Med.\ Dice} & \textbf{Prec.} & \textbf{Rec.} \\
\midrule
BUSI  ($n$=97)   & 85.3\spm{0.7} & 74.5\spm{0.3} & 90.4 & 85.5 & 83.1 \\
BRISC ($n$=860)  & 84.5\spm{0.4} & 76.7\spm{0.5} & 92.3 & 84.2 & 88.1 \\
HAM   ($n$=1503) & 94.2\spm{0.9}  & 89.8\spm{0.2} & 96.8 & 94.9 & 94.6 \\
\bottomrule
\end{tabular}
\end{table}

\section{Per-Dataset Classification Results}
\label{app:baseline_acc}

Table~\ref{tab:app_clf_summary} reports classification performance on all
three test sets. Weighted averages are reported; HAM10000 F1 slightly lower due to class imbalance (the nevus (\texttt{nv}) class constitutes 67\% of the test cases).

\begin{table}[h!]
\centering
\caption{Classification performance on test sets.}
\label{tab:app_clf_summary}
\footnotesize
\setlength{\tabcolsep}{4pt}
\begin{tabular}{lccccc}
\toprule
\textbf{Dataset} & $n$ & \textbf{Acc.} & \textbf{Prec.}
  & \textbf{Recall} & \textbf{F1} \\
\midrule
BUSI  & 117  & 93.2 \spm{0.8} & 0.93 & 0.93 & 0.93 \\
BRISC & 1000 & 99.1 \spm{0.2} & 0.99 & 0.99 & 0.99 \\
HAM   & 1503 & 87.8 \spm{0.6} & 0.87 & 0.88 & 0.87 \\
\bottomrule
\end{tabular}
\end{table}

\section{Failure Case Analysis}
\label{app:failure}

\begin{figure*}[t]
\centering
\setlength{\tabcolsep}{1.2pt}
\renewcommand{\arraystretch}{1.0}
\newlength{\fimgw}
\setlength{\fimgw}{0.1035\textwidth}  

\begin{tabular}{@{}
    ccc@{\hspace{4pt}\color{gray!60}\vrule\hspace{4pt}}
    ccc@{\hspace{4pt}\color{gray!60}\vrule\hspace{4pt}}
    ccc@{}}

\multicolumn{3}{c}{\small\textbf{BUSI}} &
\multicolumn{3}{c}{\small\textbf{BRISC}} &
\multicolumn{3}{c}{\small\textbf{HAM10000}} \\[1pt]

{\tiny Img} & {\tiny GT} & {\tiny Pred} &
{\tiny Img} & {\tiny GT} & {\tiny Pred} &
{\tiny Img} & {\tiny GT} & {\tiny Pred} \\[2pt]

\includegraphics[width=\fimgw]{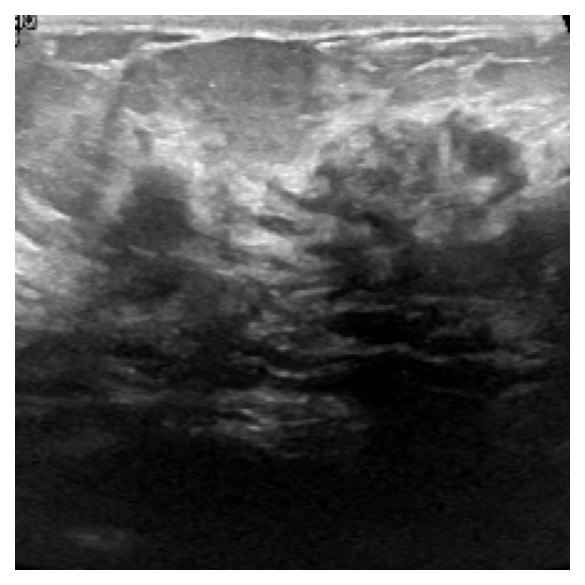} &
\includegraphics[width=\fimgw]{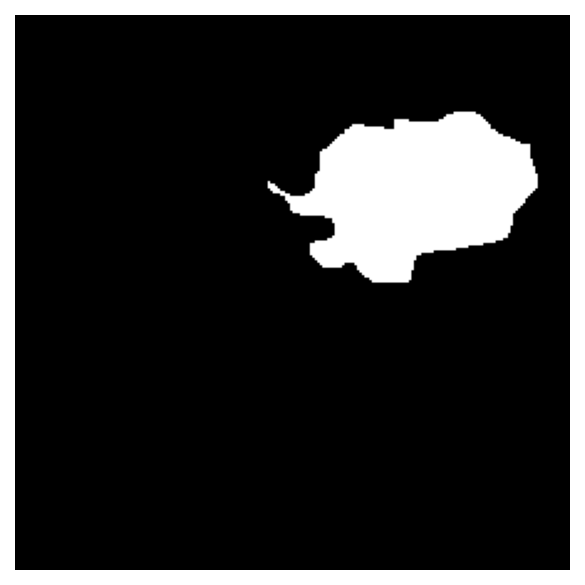} &
\includegraphics[width=\fimgw]{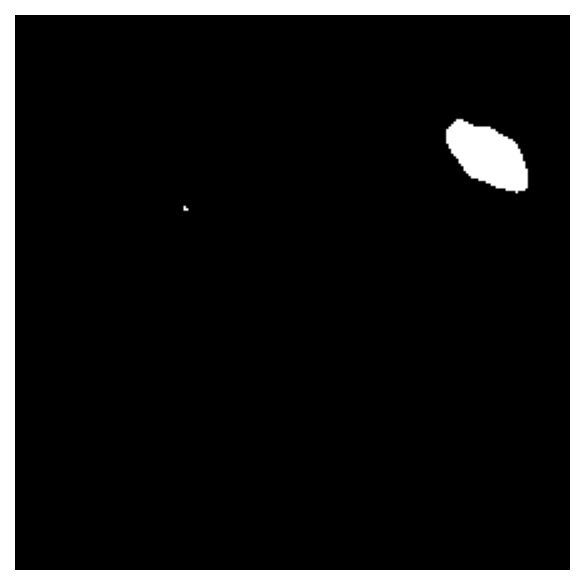} &
\includegraphics[width=\fimgw]{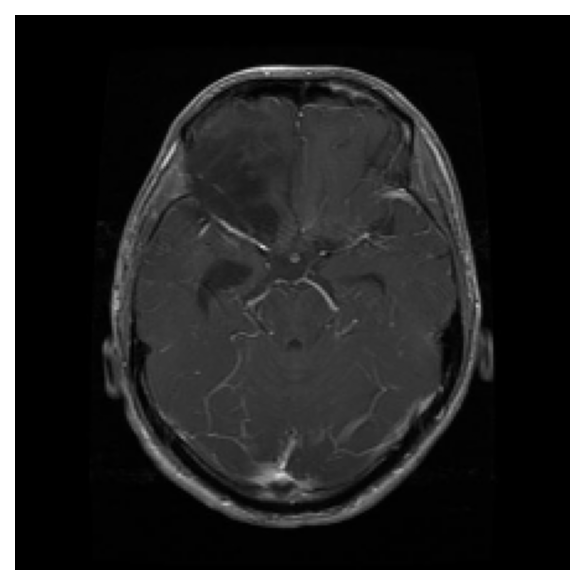} &
\includegraphics[width=\fimgw]{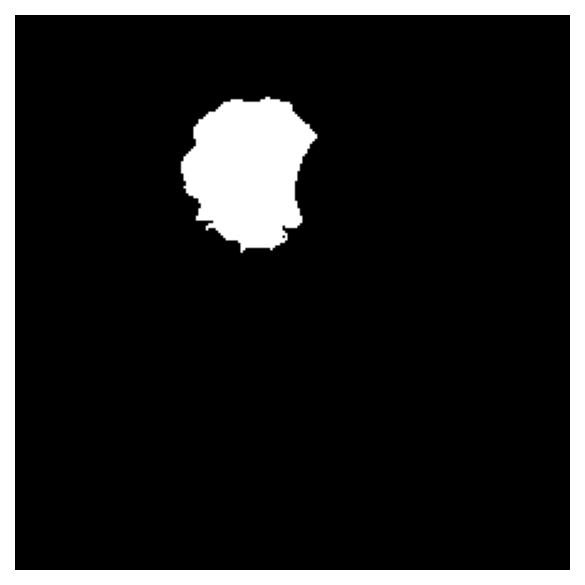} &
\includegraphics[width=\fimgw]{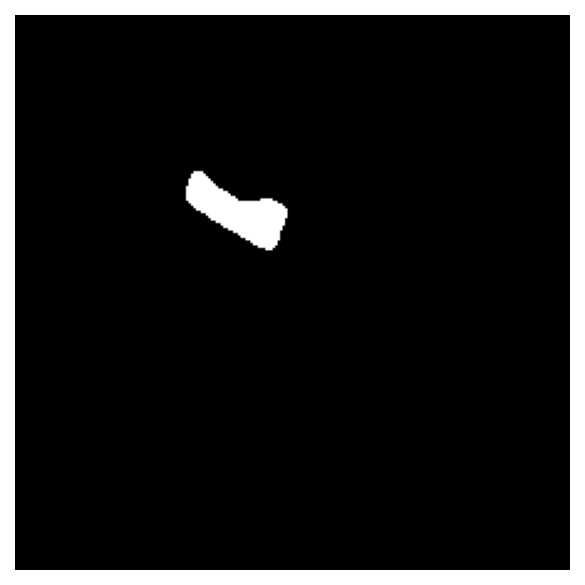} &
\includegraphics[width=\fimgw]{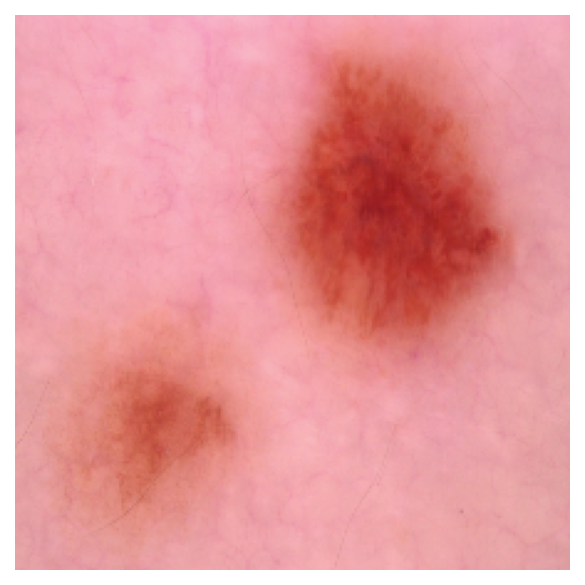} &
\includegraphics[width=\fimgw]{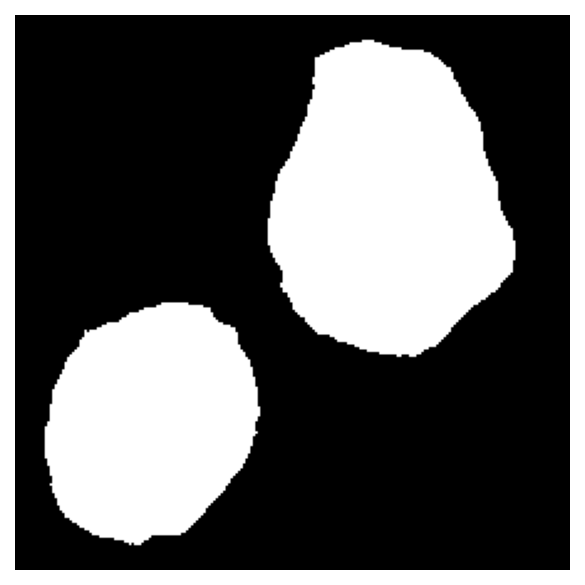} &
\includegraphics[width=\fimgw]{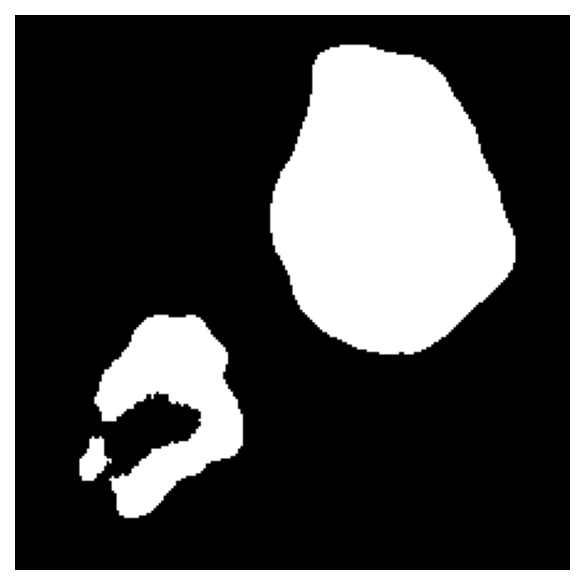} \\[1pt]

\includegraphics[width=\fimgw]{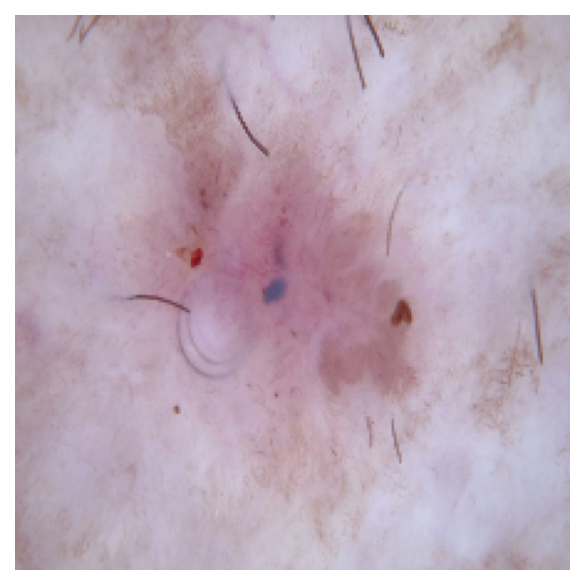} &
\includegraphics[width=\fimgw]{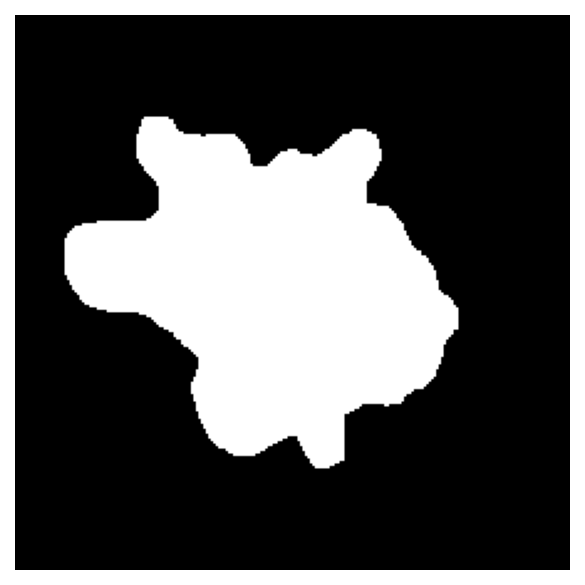} &
\includegraphics[width=\fimgw]{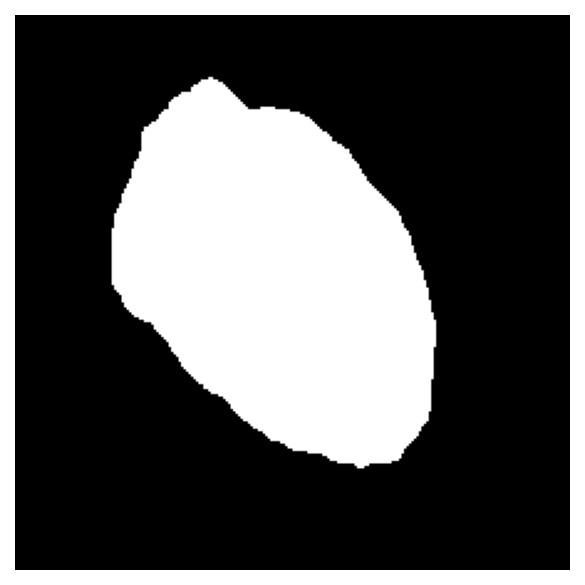} &
\includegraphics[width=\fimgw]{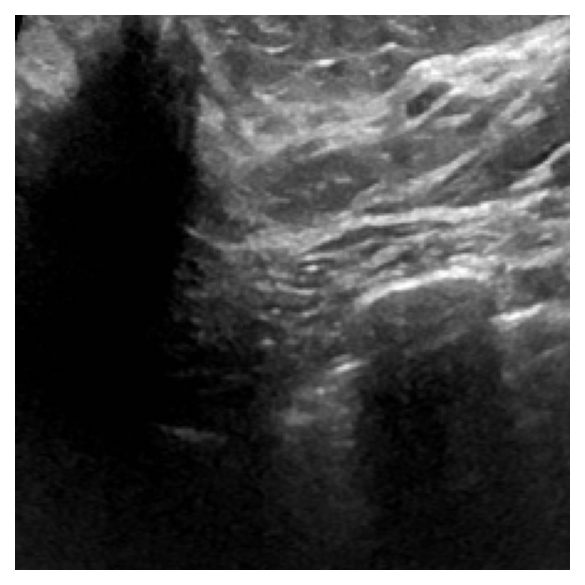} &
\includegraphics[width=\fimgw]{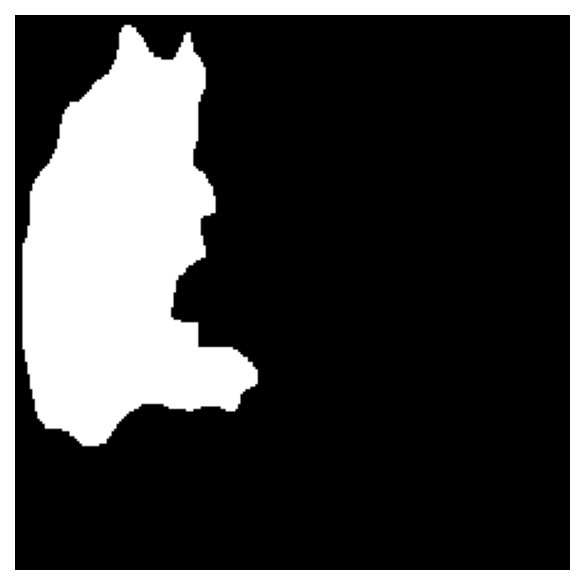} &
\includegraphics[width=\fimgw]{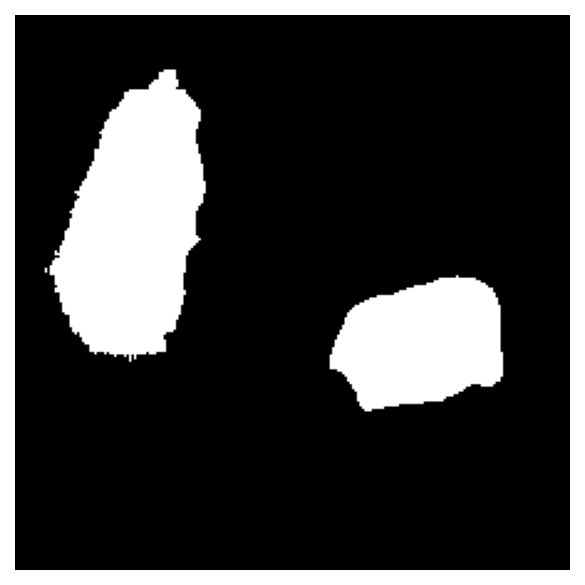} &
\includegraphics[width=\fimgw]{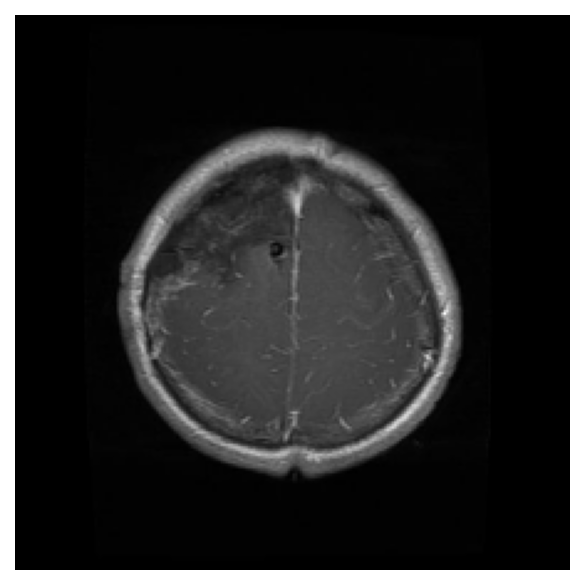} &
\includegraphics[width=\fimgw]{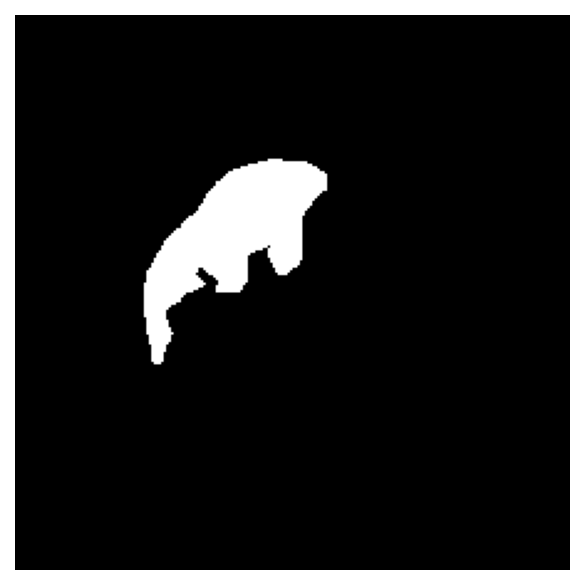} &
\includegraphics[width=\fimgw]{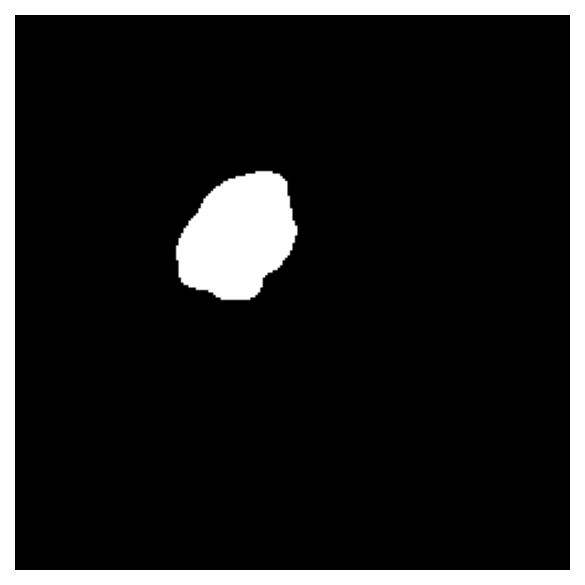} \\[1pt]

\includegraphics[width=\fimgw]{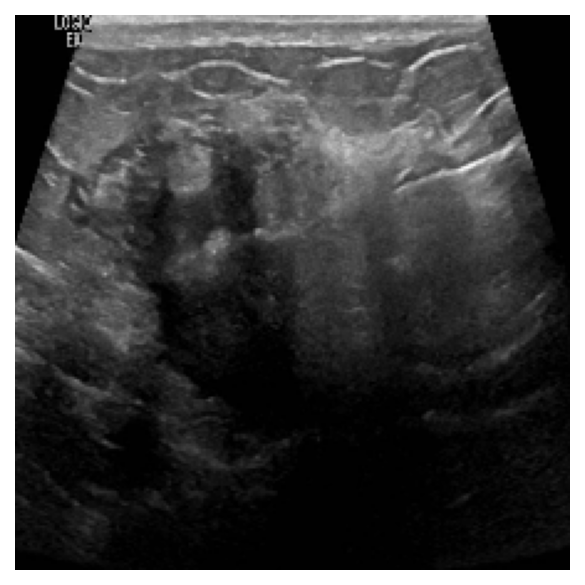} &
\includegraphics[width=\fimgw]{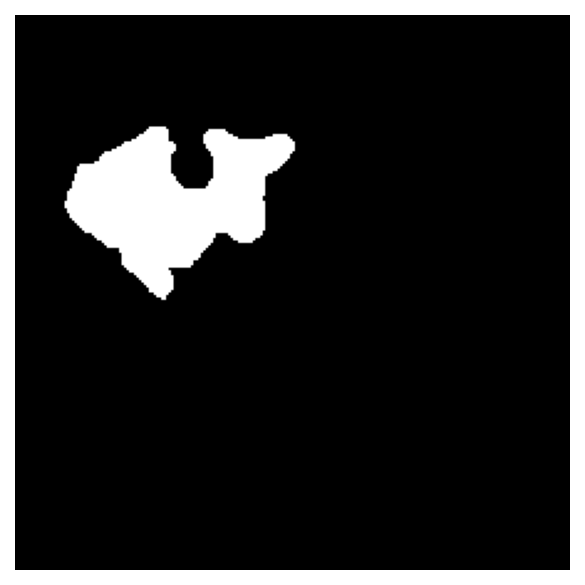} &
\includegraphics[width=\fimgw]{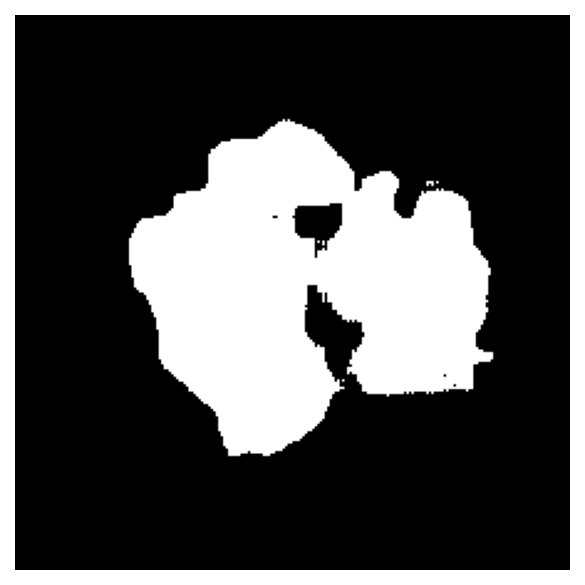} &
\includegraphics[width=\fimgw]{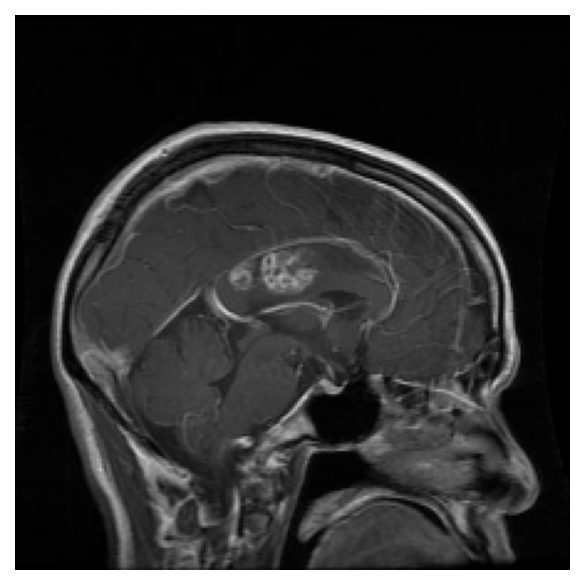} &
\includegraphics[width=\fimgw]{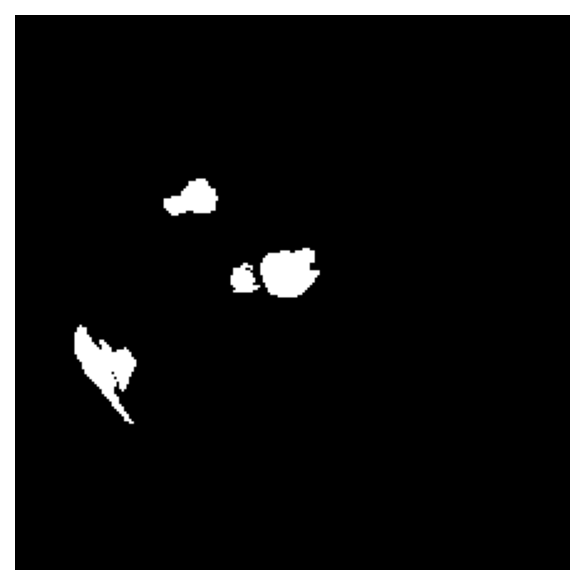} &
\includegraphics[width=\fimgw]{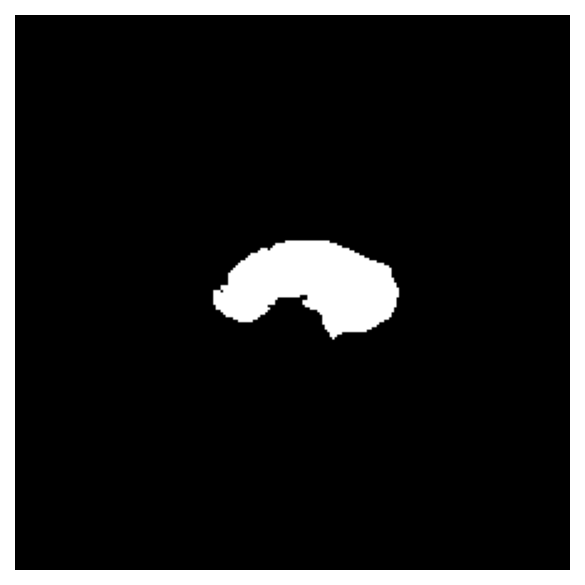} &
\includegraphics[width=\fimgw]{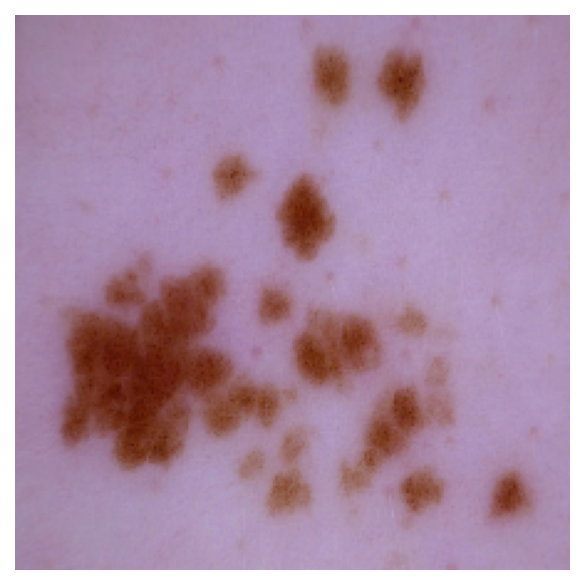} &
\includegraphics[width=\fimgw]{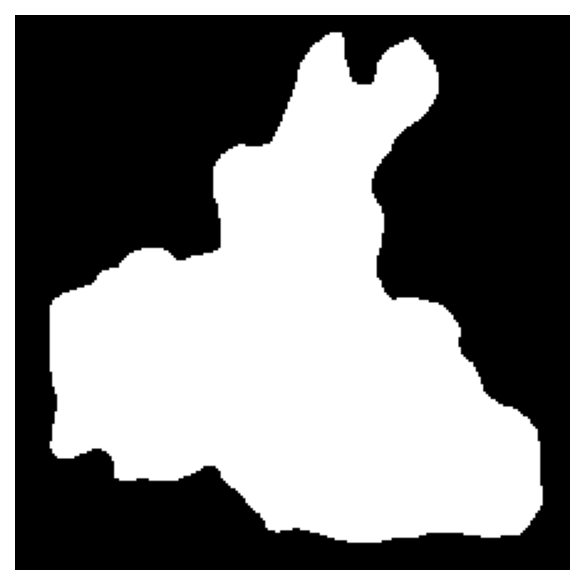} &
\includegraphics[width=\fimgw]{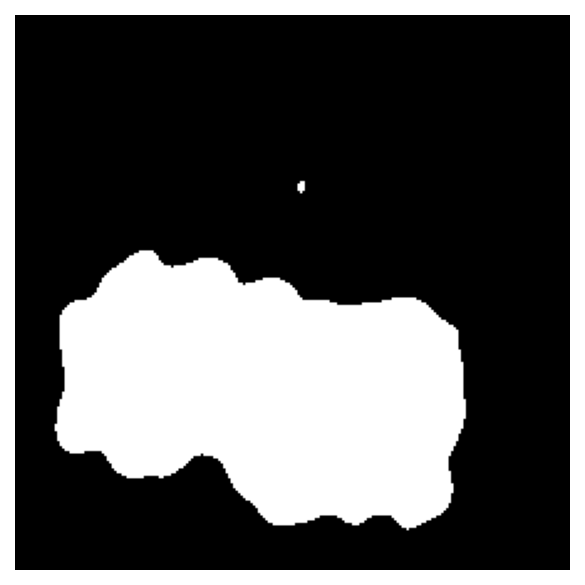} \\

\end{tabular}

\caption{Representative failure cases across all three datasets. 
Each group shows input image, ground-truth mask, and predicted mask. 
\textbf{BUSI}: acoustic shadows misidentified as lesion boundary in 
heavily shadowed regions, leading to over-segmentation. 
\textbf{BRISC}: tumour margins obscured by perilesional oedema cause 
boundary underestimation despite correct localisation. 
\textbf{HAM10000}: lesions with subtle colour gradients and ill-defined 
margins produce fragmented predictions. 
In these cases, the UPA uncertainty signals indicate unreliable 
interaction, consistent with the failure-detection analysis in Sec~\ref{app:upa}.}
\label{fig:app_failure}
\end{figure*}

Figure~\ref{fig:app_failure} presents a qualitative analysis of representative failure modes across the three imaging modalities. These comparisons illustrate how modality-specific artifacts and ambiguous pathological boundaries drive localized degradation in segmentation fidelity.

\section{Reproducibility Statement}
\label{app:repro}

All experiments use three independent random seeds (42, 123, 456)
applied to data splitting, weight initialisation, and augmentation
ordering.
The source code, including pre-processing scripts, model definitions,
training pipelines, and evaluation notebooks, will be released under an
open-source licence upon publication.
Pre-trained weights for the proposed model and all reproduced baselines,
the exact train/validation/test splits per seed, and per-seed result logs
will be made publicly available.
All experiments are reproducible from the provided configuration files
using the hardware and software environment documented in
Table~\ref{tab:app_hyper}.